\documentclass[journal]{IEEEtran}
\usepackage{ulem} 

\newcommand{\etal}{\textit{et al}.}
\newcommand{\ie}{\textit{i}.\textit{e}.}

\usepackage[misc]{ifsym}
\usepackage{algorithmicx,algorithm}
\usepackage{multirow}
\usepackage[noend]{algpseudocode}

\usepackage[pagebackref=true,breaklinks=true,linkcolor=red,anchorcolor=blue, citecolor=green,letterpaper=true,colorlinks,bookmarks=false]{hyperref}

\usepackage{breakurl}
\usepackage{graphicx}
\usepackage{amssymb}
\usepackage{bm}
\usepackage{bbm}
\usepackage{amsmath}
\usepackage{booktabs}
\usepackage{multirow}
\usepackage{makecell}
\usepackage{pifont}
\usepackage{bbding}

\usepackage{wrapfig}
\usepackage{verbatim}
\usepackage{relsize}

\usepackage{ulem}
\ifCLASSINFOpdf
\else
\fi

\begin{document}
\title{Uncertainty-Aware Distillation for Semi-Supervised Few-Shot Class-Incremental Learning}

\author{Yawen Cui, Wanxia Deng, Haoyu Chen, and Li Liu
\IEEEcompsocitemizethanks{\IEEEcompsocthanksitem This work was partially supported by National Key Research and Development Program of China No. 2021YFB3100800, the Academy of Finland under grant 331883 and the National Natural Science Foundation of China under Grant 61872379, 62022091, and the China Scholarship Council (CSC) under grant 201903170129.
\IEEEcompsocthanksitem Li Liu is with the College of System Engineering, National University of Defense Technology (NUDT), Changsha, Hunan, China. She is also with the Center for Machine Vision and Signal analysis, University of Oulu, Oulu, Finland.  Li Liu is the corresponding author.
(email:dreamliu2010@gmail.com)
\IEEEcompsocthanksitem Yawen Cui and Haoyu Chen are with CMVS, University of Oulu, Oulu, Finland. (email: yawen.cui@oulu.fi; chen.haoyu@oulu.fi)
\IEEEcompsocthanksitem Wanxia Deng is with the School of Meteorology and Oceanography, NUDT, Changsha, Hunan, China. (email: dengwanxia14@nudt.edu.cn)
}
}

\markboth{In preparation for submitting to IEEE Transactions on Neural Networks and Learning Systems}%
{Shell \MakeLowercase{\textit{et al.}}: Semi-Supervised Few-Shot Class-Incremental Learning}

\maketitle
\begin{abstract}
Given a model well-trained with a large-scale base dataset, Few-Shot Class-Incremental Learning (FSCIL) aims at incrementally learning novel classes from a few labeled samples by avoiding overfitting, without catastrophically forgetting all encountered classes previously. Currently, semi-supervised learning technique that harnesses freely-available unlabeled data to compensate for limited labeled data can boost the performance in numerous vision tasks, which heuristically can be applied to tackle issues in FSCIL, \ie, the Semi-supervised FSCIL (Semi-FSCIL). So far, \textcolor{black}{very} limited work focuses on the Semi-FSCIL task, leaving the adaptability issue of semi-supervised learning to the FSCIL task unresolved. In this paper, we focus on this adaptability issue and present a simple yet efficient Semi-FSCIL framework named Uncertainty-aware Distillation with Class-Equilibrium (UaD-CE), encompassing two modules UaD and CE. Specifically, when incorporating unlabeled data into each incremental session, we introduce the CE module that employs a class-balanced self-training to avoid the gradual dominance of easy-to-classified classes on pseudo-label generation. To distill reliable knowledge from the reference model, we further implement the UaD module that combines uncertainty-guided knowledge refinement with adaptive distillation. Comprehensive experiments on three benchmark datasets demonstrate that our method can boost the adaptability of unlabeled data with the semi-supervised learning technique in FSCIL tasks. The code is available at  \href{https://github.com/yawencui/UaD-CE}{https://github.com/yawencui/UaD-CE}. 
\end{abstract}


\IEEEpeerreviewmaketitle

\section{Introduction}
Deep learning has been successfully applied to a broad range of computer vision tasks, such as image classification \cite{krizhevsky2012imagenet,he2016deep, tang2022decision}, object detection~\cite{liu2020deep,ren2015faster,zhou2021irfr} and scene segmentation~\cite{fu2019dual, fu2020scene}, etc. However, when generalizing a trained model to unseen new classes, we need to retrain it from scratch with a large number of labeled samples together with the data from old classes. Otherwise, the discriminative ability of old classes will be undermined if we only finetune it with novel samples. Differently, humans can constantly receive and learn new concepts without forgetting old ones even with limited supervised information of new concepts, which stimulates the research interest in Few-Shot Class-Incremental Learning (FSCIL).

Based on a model well-trained by a large-scale base dataset, FSCIL~\cite{zhu2021self, tao2020few,zhang2021few, chen2021incremental, zhou2022forward, chi2022metafscil, liu2022few, chen2021semantic} aims to incrementally learn new classes in limited labeled data regime without forgetting previously seen categories. This emerging research topic faces the following challenges: (1) easy overfitting on the novel categories due to limited labeled samples; and (2) catastrophic forgetting~\cite{parisi2019continual} on old categories. Though existing FSCIL works tend to deploy intricate representation structures \cite{tao2020few,chen2021incremental} or extra feature refinement module~\cite{zhang2021few,zhu2021self} to eliminate the forgetting and overfitting issues, they still easily suffer from severe performance deterioration and classification bias problem due to the limited labeled data regime and data imbalance between the base and novel categories.

The motivation of this work relies on the intuition that utilizing freely-available unlabelled data can endue the construction of better learning procedures~\cite{van2020survey} and alleviate the above-mentioned issues in FSCIL. Our previous works~\cite{cui2021semi, cui2022uncertainty} firstly introduce semi-supervised learning into the FSCIL task by straightforwardly leveraging easily accessible unlabeled data. Although these works shed light on the Semi-Supervised FSCIL (Semi-FSCIL) solution, they only provide the baseline with existing methods and the straightforward solution, while the role of unlabeled data still cannot be maximized. Besides, there still remain some unsolved issues brought by unlabeled data, such as the high model uncertainty with the reliance on smoothness assumptions~\cite{van2020survey} and the issue of the robustness to perturbations brought by unlabeled instances. We refer to them collectively as the adaptability issue of semi-supervised learning for the FSCIL task. In this work, we focus on the adaptability issue and activate the potential value of unlabeled data to FSCIL tasks. Our experimental results argue that the adaptability issue can significantly affect the upper bound of the performance when harnessing unlabeled data in FSCIL tasks. Furthermore, the current FSCIL frameworks~\cite{zhu2021self, zhang2021few} are not compatible with the Semi-FSCIL task, since experimental results illustrate no expected remarkable performance improvement when we incorporate unlabeled data, which also emphasizes that the endowing of models' adaptability of the unlabeled data is significant.

\begin{figure}[t]
  \centering
   \includegraphics[width=1.0\linewidth]{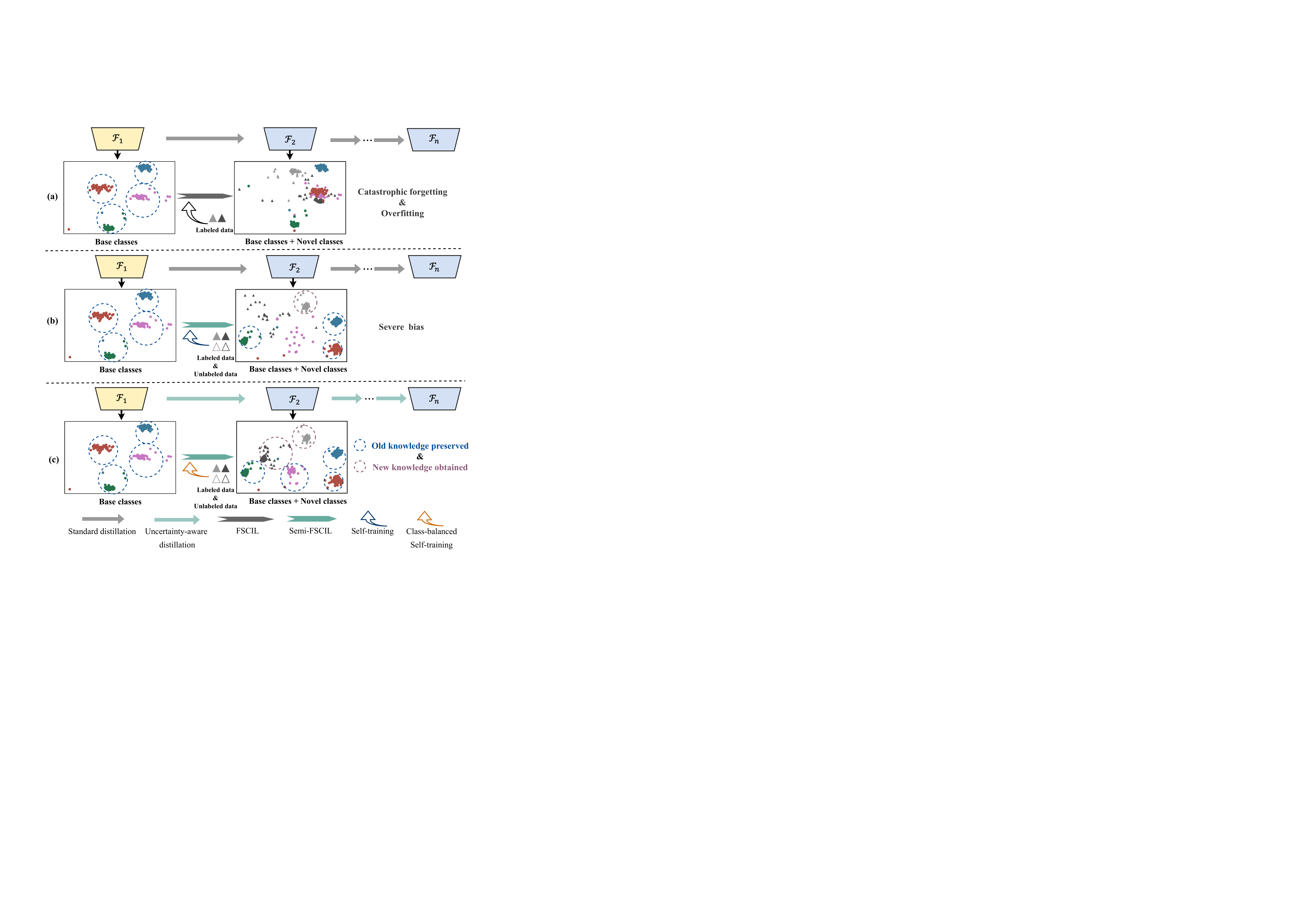}
   \vspace{-0.3in}
   \caption{\textcolor{black}{The t-SNE visualization of classification features in the FSCIL. (a) The \textbf{top}: the baseline method~\cite{rebuffi2017icarl} fails to model the distribution of novel categories with limited samples, causing severe overfitting. It also weakens the previous discriminative ability, \ie, the catastrophic forgetting of old categories. (b) The \textbf{middle}: when implementing Semi-FSCIL with standard knowledge distillation and self-training, the classifier biases to base classes with large dataset and easy-to-classified novel classes. (c) The \textbf{bottom}: with the help of unlabeled samples, our proposed framework UaD-CE can preserve the discriminative ability on base categories and obtain novel knowledge by mitigating the overfitting issue.
   }}
   \vspace{-0.1in}
   \label{fig:motivation}
\end{figure}

Toward the unresolved adaptability issues in Semi-FSCIL, we propose a simple yet efficient Semi-FSCIL framework named Uncertainty-aware Distillation with Class Equilibrium (UaD-CE). Figure~\ref{fig:motivation} describes the distribution of embedding features from our proposed method and the baseline method~\cite{rebuffi2017icarl}. As illustrated, our method can preserve the discriminative ability on previously seen categories and obtain novel knowledge by mitigating the overfitting issue concurrently. The UaD-CE encompasses two modules corresponding to the dual challenges in FSCIL: Class Equilibrium (CE) module, and Uncertainty-aware Distillation (UaD) module. 

\textcolor{black}{The data imbalance exists between the base session and the incremental session, thus the classifier always biases to base classes with a large amount of labeled data~\cite{tao2020few}. Moreover, experiments in the work~\cite{cui2022uncertainty} illustrate that there is gradual dominance of easy-to-classified novel classes on pseudo-label generation, which means that classes with a good learning status tend to be better at being recognized.} To tackle the overfitting issue on novel categories with limited labeled data and mitigate the classification bias to base categories and easy-to-classified categories, CE module is implemented with two procedures. A substantial of easily accessible unlabeled data is provided first to alleviate the data imbalance between the base and novel categories, which constitutes the Semi-FSCIL. Secondly, inspired by curriculum pseudo labeling~\cite{zhang2021flexmatch}, class-balanced self-training is employed based on the learning status of categories to avoid the gradual dominance of easy-to-classified classes on pseudo-label generation in each incremental session. The UaD module focuses on the catastrophic forgetting issue of FSCIL and alleviating the performance deterioration, where we conduct an uncertainty-guided refinement for a more efficacious exemplar set to serve the adaptive distillation process.

We naturally unify the CE module and UaD module into one single framework, which is trained end-to-end. To summarize, our main contributions include: 

\begin{itemize}

\item This paper focuses on the adaptability issue of harnessing unlabeled samples with the semi-supervised learning technique for FSCIL task, and explore the upper bound of the performance improvement with unlabeled data.


\item  To delicately conduct Semi-FSCIL, Class Equilibrium (CE) module is proposed to address overfitting and classification bias issues.

\item We introduce an efficient module named Uncertainty-aware Distillation (UaD) to distill reliable knowledge for memorizing previous categories and eliminate the ambiguity between previous and novel categories. 

\item We provide a comprehensive assessment of UaD-CE framework on three benchmark datasets to validate the effectiveness concerning three evaluation indicators.  
\end{itemize}

\section{Related Work}

\subsection{Few-Shot Learning}
Few-Shot Learning (FSL)~\cite{lai2020learning, jung2020few, yu2010attribute, vinyals2016matching,snell2017prototypical, finn2017model} aims to solve the target task with limited labeled instances per class, and there is usually a related source task whose knowledge can be transferred to the few-shot target task. Traditional FSL can be categorized into data augmentation-based methods, transfer learning-based methods, and meta-learning-based methods. Data augmentation-based methods~\cite{yu2010attribute} target at enlarging the limited labeled dataset in the instance level or the feature level.
The mechanism of transfer learning-based methods pretrain a model with a large-scale dataset first and then further finetune the model on the FSL task with the strategies of alleviating overfitting. Recent efforts on meta-learning based methods mainly follow the three directions: metric learning-based methods~\cite{vinyals2016matching,snell2017prototypical} employing metrics to evaluate the similarity among support images and query images, Optimization-based methods~\cite{finn2017model, jamal2019task} searching for better parameter configurations of the model such that it can effectively adapt to FSL tasks with a few gradient-descent update steps, and memory-based methods~\cite{santoro2016meta} aiming at forcing the query samples to match with the previously obtained knowledge. 
\subsection{Class-Incremental Learning}
Class-Incremental Learning (CIL)~\cite{rebuffi2017icarl, hou2019learning, kirkpatrick2017overcoming, zhao2021memory, zhou2021learning, liu2022model} targets at continually learning a unified classifier to recognize all seen categories so far. Great efforts have been devoted to the following two directions: identifying and preserving significant parameters of the original model \cite{kirkpatrick2017overcoming}, and memorizing the knowledge of the old model through some strategies like knowledge distillation \cite{hou2019learning}. Recently, some works focused on generalizing CIL to a limited-data regime and led to a new practical scenario, \ie, Few-Shot CIL (FSCIL) \cite{zhu2021self, tao2020few, zhang2021few, chen2021incremental, zhou2022forward, chi2022metafscil, liu2022few, chen2021semantic}. Existing methods for FSCIL mainly employ two strategies. The one is knowledge representation and refinement. Tao \etal~\cite{tao2020few} propose TOPIC to model the topology of the feature space using neural gas. Zhang \etal~\cite{zhang2021few} adopt a simple but effective decoupled learning strategy of representations, and Continually Evolved Classifier (CEC) is proposed by employing a graph model to propagate context information between classifiers for adaptation. Zhu \etal ~\cite{zhu2021self} offer a novel incremental prototype learning scheme to solve the FSCIL task. Another strategy is via knowledge distillation. Cheraghian \etal ~\cite{cheraghian2021semantic} employ the semantic information during training, and an attention mechanism on multiple parallel embeddings of visual data is proposed to align visual and semantic vectors, which reduces issues related to catastrophic forgetting. 
Recently, we benchmark the Semi-FSCIL task first and provide a detailed configuration and straightforward solution~\cite{cui2021semi}. Then, we further consider the uncertainty in semi-supervised learning process to promote the performance, while the adaptability issues can not be solved better since the gain obtained with unlabeled data is limited and the performance imbalance among categories is severe~\cite{cui2022uncertainty}.
In this paper, we still focus on the unresolved adaptive issues of Semi-FSCIL task and discuss the effect of uncertainty on knowledge distillation procedure.

\subsection{Semi-Supervised Learning}
Semi-Supervised Learning~\cite{song2022graph, berthelot2019mixmatch, lee2013pseudo, sohn2020fixmatch, iscen2019label} is to promote the supervised learning performance with unlabeled samples. Diverse methods focus on semi-supervised learning, which can be divided into consistency-regularization methods~\cite{berthelot2019mixmatch, sohn2020fixmatch, zhang2021flexmatch} and  pseudo-labeling~\cite{lee2013pseudo, rizve2021defense}. Consistency-regularization methods always rely on an extensive set of data augmentations requiring domain-specific knowledge that is insufficient for the limited-data regime (one or few labeled examples per category). FixMatch~\cite{sohn2020fixmatch} generates the pseudo-label for a weakly-augmented unlabeled image first, then the model is trained to predict the pseudo-label of a strongly-augmented version of the same image. Pseudo-labeling methods aim to assign pseudo labels for unlabeled data by using the model trained on labeled ones. Specifically, the pseudo labels can be created by the predictions of trained neural network \cite{lee2013pseudo, rizve2021defense} or assigned based on neighborhood graph \cite{iscen2019label, zhu2003semi, liu2019deep}. Zhu~\etal~\cite{zhu2003semi} represent labeled and unlabeled data as vertices in a weighted graph with edge weights encoding the similarity between instances. In this paper, we employ class-balanced self-training to avoid the bias to easy-to-classified classes on pseudo-label generation.

\subsection{Uncertainty Estimation}
There are mainly two types of uncertainty \cite{kendall2017uncertainties}: aleatoric uncertainty which models the noise inherent in the data, and epistemic uncertainty which accounts for the uncertainty in the model. Aleatoric uncertainty can not be lessened even if we collect more data, while epistemic uncertainty can be explained away when enough data is provided. In this paper, we mainly focus on epistemic uncertainty. Traditionally, approximate Bayesian inference methods~\cite{louizos2016structured} can be applied to obtain the model uncertainties. However, due to the costly computation and the implementation hardness of Bayesian neural network, Gal~\etal~\cite{gal2016dropout} prove that dropout together with its variants can be seemed as a Bayesian approximation to represent model uncertainty in deep learning. The test time data augmentation~\cite{wang2018automatic,gawlikowski2021survey} is one of the simpler
predictive uncertainty estimation techniques, which are also applied in this paper.

\begin{figure*}[t]
  \centering

   \includegraphics[width=1.0\linewidth]{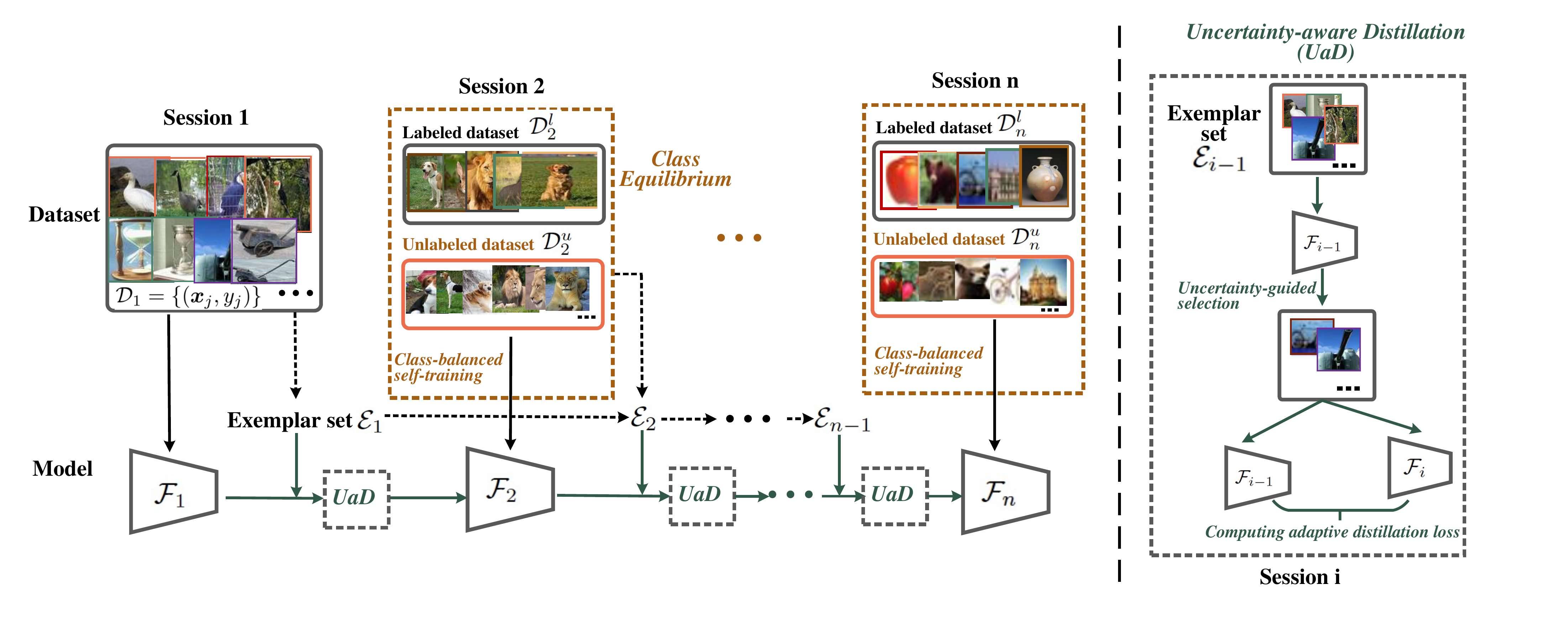}
   \vspace{-0.05in}
   \caption{Illustration of the proposed UaD-CE framework. First, the model is trained on the large-scale $\mathcal{D}_{1}$. When it comes to the following sessions, we incorporate the unlabeled set into the labeled training set to alleviate the inadequate of novel samples and propose to utilize the class-balanced self-training as the semi-supervised learning method. In order to solve the catastrophic forgetting problem efficiently, the proposed UaD module is conducted by eliminating the negative effect of unlabeled samples.}
   \label{fig:method}
   \vspace{-0.05in}
\end{figure*}

\section{Methodology}
In this section, we introduce the semi-supervised learning framework UaD-CE for Semi-FSCIL as shown in Fig. \ref{fig:method}. Our UaD-CE framework contains two key components: Class Equilibrium (CE) module for tackling the overfitting issue with large-scale unlabeled samples via semi-supervised learning, and Uncertainty-aware Distillation (UaD) module for distilling reliable knowledge from large-scale yet unlabeled samples and eliminating the ambiguity between the current categories and previous categories. We first give the problem formation with learning targets, then the overview of the framework, and lastly the details of each component in the framework will be introduced.
\subsection{Problem Formulation}
We present a sequence of disjoint datasets by $\mathcal{D}=\{\mathcal{D}_1, \mathcal{D}_2, ...,\mathcal{D}_n\}$, where $\mathcal{D}_{1}$ is the large-scale base dataset used in the first base session and the followings are all novel few-shot datasets. To be specific, we define $\mathcal{D}_{1}=\left\{\left(\bm{x}_j, y_j\right)\right\}_{j=1}^{|\mathcal{D}_1|}$ where $ y_j \in \mathcal{C}_1$ and $\mathcal{C}_1$ denotes the base category set. In the $i$-th session where $i>1$, the novel/new category dataset is set as $\mathcal{D}_{i}=\mathcal{D}_{i}^l \cup \mathcal{D}_{i}^u$ with $\mathcal{D}_{i}^l$ and $\mathcal{D}_{i}^u$ denoting the labeled training data and unlabeled training data, respectively. The labeled training data $\mathcal{D}_{i}^l=\left\{\left(\bm{x}_j, y_j\right)\right\}^{N \times K}_{j=1}$ consists of $N$ classes $\mathcal{C}_i$ with $K$ labeled examples per class, \ie, a $N$-way $K$-shot problem. The unlabeled training data $\mathcal{D}_{i}^u=\left\{\bm{x}_j\right\}^{M}_{j=1}$ comprises unlabeled samples, where $M\gg K$. $j$ is the index of a specific sample in $\mathcal{D}^u$ and $\mathcal{D}^l$. Noteworthy, there is no overlap between the categories of different sessions, \ie, $\mathcal{C}_{i} \cap \mathcal{C}_{i'} = \emptyset$, where $i\neq i'$. In this paper, we implement and validate our proposed method on object classification task. As for the model $\mathcal{F}\left(\cdot\right)$, it usually contains the backbone $\Theta\left(\cdot\right)$ for extracting features and the classification head $\Gamma\left(\cdot\right)$. 

The aim of Semi-FSCIL is to train a unified classification model with a sequence of disjoint datasets delicately, and it requires the model to incrementally learn the new categories without forgetting all the learned categories so far, \ie, in session $i$, after being trained on the $\mathcal{D}_i$,  $\mathcal{F}\left(\cdot\right)$ needs to classify the samples from categories of $\mathcal{C}_1 \cup \mathcal{C}_2 \cup, ..., \mathcal{C}_i$. 

\subsection{Overview of UaD-CE Framework}

Our UaD-CE framework arranged by data flow is illustrated in Figure.~\ref{fig:method}. First, $\mathcal{F}_1$ is trained with base dataset $\mathcal{D}_{1}$ by computing the classification loss. For the sake of preserving the current classification performance later, an exemplar set $\mathcal{E}_1$ filtered from $\mathcal{D}_{1}$ is stored in an extra memory. When it comes to the second session, $\mathcal{F}_2$ is initialized by the reference model $\mathcal{F}_{1}$, \ie, the teacher model. Then, class-balanced self-training is conducted with $\mathcal{D}_{2}^{l}$ and $\mathcal{D}_{2}^{u}$ for learning novel categories by mitigating the overfitting issue, which is the Class-Equilibrium (CE) module. While for the categories encountered in the previous training, Uncertainty-aware Distillation (UaD) is implemented by distilling reliable knowledge from the reference with the exemplar set $\mathcal{E}_1$. At the end of the current session, $\mathcal{E}_1$ is updated into $\mathcal{E}_2$ by adding the filtered samples from the current training dataset. The procedure of the second session will be iteratively executed for the rest sessions to accomplish the Semi-FSCIL task. The pseudo code of our proposed methods is shown in Algorithm~\ref{Algo:the process}.

\begin{algorithm}[t]

\caption{Semi-FSCIL with \textbf{UaD-CE}.}
\hspace*{0.02in} {\bf Input:} 
{$\mathcal{D}_{1}$, $\mathcal{D}_{2}$, ..., $\mathcal{F}\left(\cdot\right)$, session number $n$.}\\
\hspace*{0.02in} {\bf Output:}$\mathcal{F}\left(\cdot\right)$ that can classify all seen categories so far.
\begin{algorithmic}[1]
\For{$i$ in $n$} 
\If{$n$==1} 
\State Train $\Theta\left(\cdot\right)$, $\Gamma\left(\cdot\right)$ by  $\mathcal{D}^{l}_1$ to form $\mathcal{F}_1$; 
\State Sample exemplars $\mathcal{E}_1$ from $\mathcal{D}^{1}$;
\Else
\State $\mathcal{F}_{ref}$ = $\mathcal{F}_{i-1}$;
\State $\mathcal{F}_{target}$ = $\mathcal{F}_{ref}$;
\For{supervised epochs}
\State $\mathcal{F}_{target}$ updates by learning $\mathcal{E} \cup \mathcal{D}^{l}_i$;
\EndFor
\For {unlabeled iterations}
\State Train $\mathcal{F}_{target}$ on ${\mathcal{D}}^{l}_i$ and ${\mathcal{D}}^{u}_i$ by \textbf{class-balanced self-training};
\State Do \textbf{uncertainty-aware distillation} with $\mathcal{F}_{ref}$ and $\mathcal{F}_{target}$;
\EndFor
\State Update $\mathcal{E}_{i-1}$ to $\mathcal{E}_{i}$ by sampling from ${\mathcal{D}}^{l}_i$ and ${\mathcal{D}}^{u}_i$;
\EndIf
\State $\mathcal{F}_i$ = $\mathcal{F}_{target}$;
\State Conduct the classification by NME;
\EndFor
\State
\Return $\mathcal{F}\left(\cdot\right)$ obtained after $n$ sessions.
\end{algorithmic}
\label{Algo:the process}
\end{algorithm}

\vspace{-0.1in}
\subsection{Class Equilibrium}
Through the evaluation of existing methods~\cite{zhang2021few,zhu2021self} for FSCIL, we find that their overall performances are mainly contributed by the high classification accuracy of base categories that possess massive labeled samples. However, the overfitting issue of novel categories is still severe due to the few-shot learning regime, which causes the huge performance discrepancy between base and novel categories. Semi-FSCIL~\cite{cui2021semi} can resolve this bias issue mainly caused by the imbalance among datasets by incorporating a freely-used unlabeled dataset in each incremental session, which is also adopted in CE module first.

When harnessing unlabeled data, this work~\cite{cui2021semi} utilizes the self-training~\cite{grandvalet2004semi} as the semi-supervised learning method. Since self-training selects pseudo labels based on the prediction probability with the smoothness assumptions~\cite{van2020survey}, it tends to bias to easy-to-classified classes, ignoring other classes that have inferior classification performances. In this paper, we take this bias and the learning status into consideration and apply the class-balanced self-training for Semi-FSCIL which is the critical component of CE module.

The CE module is conducted from the second session since a large-scale dataset is available in the first session. As shown in Figure~\ref{fig:method}, in session $i$ ($i \textgreater 1$), an unlabeled dataset $\mathcal{D}_i^u=\left\{\bm{x}_j\right\}^{M}_{j=1}$ is introduced into each incremental session to alleviate the class-imbalance between the base and novel categories first. In this way, the model can learn these novel categories in a semi-supervised manner. The common self-training process contains two procedures: supervised epochs and unlabeled iteration. In supervised epochs, the model is trained with labeled samples first. After that, unlabeled iterations are conducted $L$ times for selecting unlabeled samples together with obtained pseudo labels from the current $\mathcal{F}$ into the training process. For each unlabeled iteration, the obtained pseudo labels are termed $\widetilde{Y}_i=\left\{\widetilde{y}_j\right\}^{M}_{j=1}$, in which $\widetilde{y}_j$ obtained by the following: 

\begin{equation}
\widetilde{y}_j = \mathbbmss {1} \left[p_j^c == \bm{{\rm max}}(\bm{p}(\bm{x}_j))\right]_{c=1}^{N},
\label{eq:y}
\end{equation}
where $p_j^c$ is the probability that this unlabeled sample belongs to the $c^{th}$ category of the total $N$ categories in session $i$, and $\bm{{\rm max}}(\bm{p}(\bm{x}_j))$ is the maximum of $\bm{p}(\bm{x}_j)$.

Typically, a threshold $\gamma$ for $\bm{{\rm max}}(\bm{p}(\bm{x}))$ is defined for selecting the unlabeled samples $\widehat{\mathcal{D}}_{i,t}^u$ together with their pseudo labels. Inspired by curriculum pseudo labeling~\cite{zhang2021flexmatch}, the learning status of a specific class can be reflected by the numbers of unlabeled samples whose predictions fall into this class and are above $\gamma$. Classes that are well-learned by the model or easy-to-classified tends to have higher prediction probability. In this way, these classes do not require overmuch unlabeled data. To avoid the gradual dominance of easy-to-classified classes on pseudo-label generation, we propose to select unlabeled samples from class level. 

First, we divide the $\widehat{\mathcal{D}}_{i,t}^u$ based on the acquired pseudo labels, and for each specific category, we obtain a subset of unlabeled dataset $\mathcal{D}_{i,t}^{u} = \{(\bm{x}_j, \widetilde{y}_j)\}$ representing the unlabeled dataset assigned the pseudo label by the $t^{th}$ category in session $i$. Due to the $N$-way $K$-shot problem in each incremental session, we will obtain $N$ subsets after the division process. When choosing unlabeled data with the consideration of the learning status, we introduce a separate parameter $p_{i,t}$ determining the proportion of selected
instances for a specific category. Consequently, the selected unlabeled set in each unlabeled iteration is
\begin{equation}
\widetilde{\mathcal{D}}_{i}^u = \{\widetilde{\mathcal{D}}_{i,t}^u | t=1,2, ...,N\},
\end{equation}
where $\widetilde{\mathcal{D}}_{i,t}^u$ is made of the first  $p_{i,t}$ partition of $\widehat{\mathcal{D}}_{i,t}^u$ ranked by prediction probabilities. The setting of $p$ should fulfill that 
\begin{equation}
\forall \; |\widehat{\mathcal{D}}_{i,a}^u| \leq |\widehat{\mathcal{D}}_{i,b}^u|,\; p_{i,a} \geq p_{i,b},
\end{equation}
where $a$ and $b$ stand for the $a^{th}$ and $b^{th}$ categories in session $i$. This restriction guarantees that more unlabeled samples for hard-to-classified categories are incorporated into the training process.

\subsection{Uncertainty-Aware Distillation} 
Another challenge of FSCIL is catastrophic forgetting on previously seen categories. The distillation-based framework is an effective solution for maintaining previous abilities by transferring the related knowledge from the reference model to the target model, which is verified in large-scale CIL tasks. This work~\cite{cui2021semi} for Semi-FSCIL also learns from knowledge distillation technique by simply implementing this task on existing CIL methods~\cite{hou2019learning,rebuffi2017icarl}. However, The work~\cite{cui2021semi} ignores that the challenges faced by large-scale CIL tasks and FSCIL tasks are not exactly the same, and the high model uncertainty with the combination of unlabeled samples results in the unstable distillation process. In this section, we introduce the Uncertainty-aware Distillation module to solve the issues in the distillation-based framework for Semi-FSCIL. 

The mechanism of the standard knowledge distillation technique is to evaluate prediction variations on old categories after the model is updated by new categories of the current session. Typically, a distillation loss is introduced to the standard classification loss. Distillation loss occurs between two models used in a pair of neighboring sessions. The model in the previous session is termed the reference model, and another one in the current session is the target model. Apart from the preserved reference model, part of old class samples, termed exemplars, are required to store in an extra memory for access to the current session. To this end, the incremental learning loss function in a specific session $i$ is defined as follows:
\begin{equation}
\mathcal{L}(\mathcal{D}_i,\mathcal{E}_i, \mathcal{F}) = \mathcal{L}_{ce}(\mathcal{D}_i,\mathcal{E}_i, \mathcal{F}) + \mathcal{L}_{dl}(\mathcal{E}_i, \mathcal{F}),
\label{eq:loss_total_old}
\end{equation}
where $\mathcal{L}_{ce}$ means the cross-entropy loss and $\mathcal{L}_{dl}$ denotes the distillation loss, and $\mathcal{E}_i$ is old class exemplars drawn from the datasets in previous $i-1$ sessions. 

Due to unlabeled data being introduced into the incremental learning session, the extra unlabeled data may be preserved as the exemplars and participate knowledge distillation process in the following sessions. Precisely, $\mathcal{E}_i$ contains labeled samples and unlabeled samples at the same time. 
According to Equation~\ref{eq:loss_total_old}, the reference model generates objectives for the target model, \ie, the reference model makes predictions on exemplars. However, the predictions on exemplars from the unlabeled dataset are uncertain because of the unstable training process with limited labeled samples. Accordingly, the distillation loss is unreliable for memorizing the previous-seen categories. To transfer trustworthy knowledge from the reference model, we propose an Uncertainty-aware Distillation for the Semi-FSCIL configuration.

The proposed UaD contains two components: uncertainty-guided refinement and adaptive distillation loss. To promote the efficacy of knowledge distillation, the uncertainty-guided refinement acts on the exemplar set to obtain more reliable instances with more certain objectives obtained with the reference model. First, the uncertainty of predictions on the exemplar set ~\cite{gawlikowski2021survey} is estimated with data augmentation. Assuming that the uncertainty is termed $\lambda_{i,q}$ for the $q^{th}$ exemplar in session $i$, the updated exemplar set is defined as 

\begin{equation}
\widetilde{\mathcal{E}}_i = \mathcal{E}_i \left[ \mathbbmss {1} \left[\lambda_{i,q} > \tau \right]_{q=0}^{|\mathcal{E}_i|} \right],
\label{eq:update-exemplar}
\end{equation}
where $\tau$ is the uncertainty threshold. After the refinement, exemplars with stable predictions are remained, which represents that old knowledge in the reference model can be effectively transferred by computing losses on reliable exemplars. 

Considering the quality of distilled knowledge, the refinement is conducted to remove uncertain exemplars. With the quantity concern, we define the adaptive distillation procedure with a dynamically changeable weight $\zeta_i$:
\begin{equation}
\zeta_i = \zeta^{base} \times |\mathcal{E}_i|/|\widetilde{\mathcal{E}}_i| \times \sqrt{|C^{old}_i|/|C^{new}_i|},
\label{eq:adapt}
\end{equation}
where $\zeta^{base}$ is a fixed term decided by the dataset. $|\mathcal{E}_i|/|\widetilde{\mathcal{E}}_i|$ adjusts the amount of distilled old knowledge within a particular incremental session. $|C^{old}_i|$ and $|C^{new}_i|$ are the numbers of seen categories previously and novel encountered categories in session $i$. $\sqrt{|C^{old}_i|/|C^{new}_i|}$ considers the ratio of the previous class number and the novel class number in the current session, which is also appeared in \cite{hou2019learning}. To this end, Equation~\ref{eq:loss_total_old} is redefined as 
\begin{equation}
\mathcal{L}(\mathcal{D}_i,\widetilde{\mathcal{E}}_i, \mathcal{F}) = \mathcal{L}_{ce}(\mathcal{D}_i,\widetilde{\mathcal{E}}_i, \mathcal{F}) +  \zeta_i \mathcal{L}_{dl}(\widetilde{\mathcal{E}}_i, \mathcal{F}),
\label{eq:loss_total}
\end{equation}
where we use $\mathcal{F}$ to represent the model set. When computing distillation loss, the $\mathcal{F}$ contains the reference model and the target model. The modified distillation loss ensures that the old knowledge can be memorized efficiently and balanced with new knowledge to eliminate the ambiguity between old and novel categories in the overall classification process.

\section{Experiments}
\subsection{Dataset and Evaluation Indicators}
We conducted comprehensive experiments on three benchmark datasets for FSCIL: CIFAR100~\cite{krizhevsky2009learning}, \textit{mini}ImageNet~\cite{vinyals2016matching} and CUB200~\cite{chaudhry2018efficient}. The dataset configurations are illustrated in Table \ref{table:dataset}. \textbf{CIFAR100}~\cite{krizhevsky2009learning} is commonly used in CIL. This dataset concludes $100$ categories with $600$ RGB images per class. For each category, $500$ images are used for training and 100 images for testing. The size of the image is $32 \times 32$.  \textbf{\textit{mini}ImageNet}~\cite{vinyals2016matching} is a subset of the ImageNet with a smaller number of classes. It includes $600$ images for each of $100$ classes. These images are of the size of $84 \times 84$. \textbf{CUB200}~\cite{chaudhry2018efficient} contains about $6,000$ training images and $6,000$ test images of over $200$ bird categories. The images are resized
to $256 \times 256$ and then cropped to $224 \times 224$ for training. 
By following the dataset configuration in \cite{tao2020few}, the dataset configuration for FSCIL is illustrated in Table~\ref{table:dataset}. For CIFAR100 and \textit{mini}ImageNet, we set 60 and 40 classes as the base and novel categories, respectively, and chose a 5-way 5-shot setting in each incremental session. In total, we had 9 training sessions, \ie one session for base classes and 8 sessions for novel classes.
While for CUB200, we chose 100 classes as base classes and split the remaining 100 classes into 10 incremental sessions with the 10-way 5-shot setting. Notably, except for labeled samples used in each incremental learning session, the rest was regarded as unlabeled dataset, which is followed by the work~\cite{cui2021semi}.

\begin{table}[]
\caption{The dataset configurations for FSCIL. \#Categories and \#Samples stand for the number of categories and the number of samples, respectively. The learning pattern represents the setting of novel tasks in each incremental learning session.}
\centering
\setlength{\tabcolsep}{11pt}
\resizebox{8.8cm}{!}{
\tabcolsep 0.06in
\begin{tabular}{ccc|ccc}
\bottomrule[1.3pt]
    & \multicolumn{2}{c|}{\textbf{Base session}} & \multicolumn{3}{c}{\textbf{Incremental session}} \\ \cline{2-6} 
                         &    \#Categories       &   \#Samples  &   \#Categories    &   \#Samples    &   Incremental pattern   \\ \cline{2-6} 
 \multirow{1}{*}{\textbf{CIFAR100} }                             &     60      &     500      & 40      &    5   &   5-way 5-shot   \\ 
 \multirow{1}{*}{\textbf{\textit{mini}ImageNet} }                             &    60       & 500      &   40    &    5   &  5-way 5-shot    \\ 
  \multirow{1}{*}{\textbf{CUB200} }   &                   100       &       30    &     100     &   5    &   10-way 5-shot         \\                             
\bottomrule[1.3pt]
\end{tabular}
}
\label{table:dataset}
\end{table}

We conducted on three evaluation indicators: (1) the final overall accuracy ($\%$) in the last session; (2) the performance dropping rate (PD) ($\%$) \cite{zhang2021few} that measures
the absolute accuracy drops in the last session w.r.t. the accuracy in the first session; (3) the average accuracy ($\%$) of all the sessions. 

\subsection{Model Configurations} In our experiment, ResNet-18~\cite{he2016deep} was employed as the backbone for CIFAR100, \textit{mini}ImageNet and CUB200. As for the backbone, we froze the parameters of the front four layers after the first session. During training, the model was optimized by SGD \cite{robbins1951stochastic} (with lr=0.1 and wd=5e-4). When selecting exemplar set incorporated into the following sessions for memorizing the previous session categories by knowledge distillation, we used the method proposed in \cite{rebuffi2017icarl} based on herding selection. Moreover, we applied the nearest-mean-of-exemplars classification, donated as NME. The whole framework was implemented using Pytorch and trained on GeForce RTX 3080 GPUs. 

\subsection{Parameter Configurations and Training Details} 
For the first session of CIFAR100 and \textit{mini}ImageNet, the learning rate started from 0.1 and was divided by 10 after 80 and 120 epochs (160 epochs in total). For the rest sessions, the learning rate was 0.001 in 100 epochs. For CUB200, the base learning rate in the first session was 0.001, and divided by 10 after 80 and 120 epochs (160 epochs in total). The learning rate of the following sessions was 0.0005 used in a total of 60 supervised epochs. The model was trained with the training batch size of 128 for \textit{mini}ImageNet, and 32 for CIFAR100 and CUB200. The test batch size was 100 for \textit{mini}ImageNet, and 50 for CIFAR100 and CUB200. 

For each unlabeled iteration, we chose 10 unlabeled samples incorporated into the training procedure with a class-balanced manner. As for the the selection proportion $p$, this parameter ensures that the numbers of unlabeled samples added to each class are equal. In each unlabeled iteration, the model was trained 
on the labeled dataset and chosen unlabeled dataset with pseudo labels for extra 10 epochs for CIFAR100 and \textit{mini}ImageNet, 20 epochs for CUB200. 

To guarantee that labeled samples contributed more to the training process, the model was not trained from the reference model with more epochs in each unlabeled iteration, instead continually trained after labeled epochs. Moreover, during the extra epochs, the training dataset was not set randomly to ensure that the model was trained on labeled data first, and losses could be generated again from unlabeled samples. Finally, when computing the class means, the dimension was set as 512.

For CIFAR100, we added 350 unlabeled samples in 35 unlabeled iterations, while for \textit{mini}ImageNet and CUB200, we incorporated 160 unlabeled samples in 16 unlabeled iterations. In the uncertainty-guided selection part, we added random Gaussian noise to the input for obtaining the model uncertainty. For each sample, 10 forward passes with random Gaussian noise were conducted, and the variance on predictions was regarded as the uncertainty. Then, we ranked the exemplars in an ascending order based on the uncertainty. Notably, we selected the top three-quarter exemplars which then were used for computing the adaptive distillation loss. In the adaptive weight, $\zeta^{base}$ was assigned 1 to CIFAR100, 2 to \textit{mini}ImageNet and CUB200. As for the exemplar set, we selected 20 samples per class of previous sessions. The number is equal to that of large-scale CIL setting~\cite{hou2019learning}.

\begin{figure*}[t]
  \centering
   \includegraphics[width=0.9\linewidth]{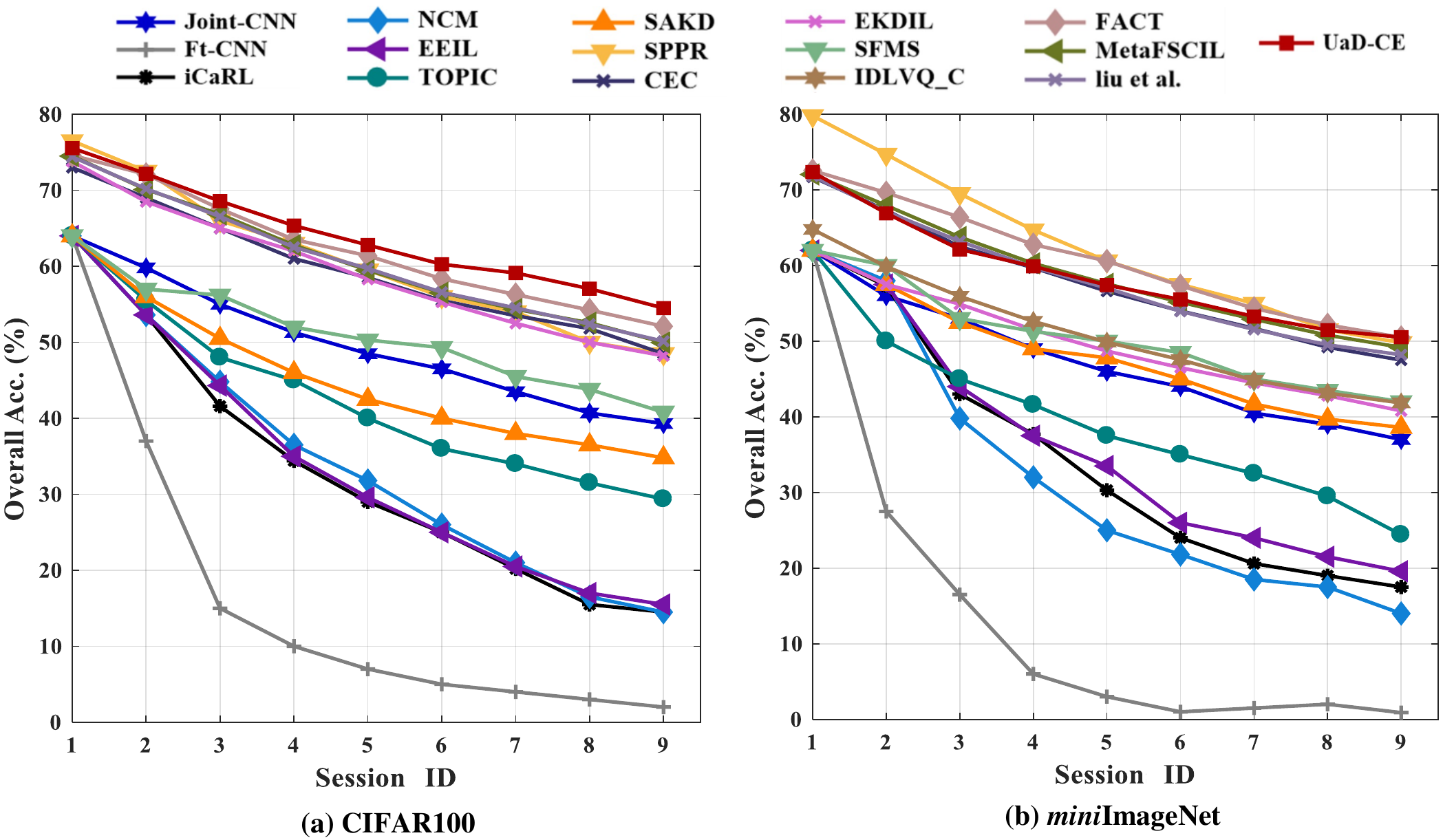}
   \vspace{-0.10in}
   \caption{Comparative study on the FSCIL task with CIFAR100 (a) and \textit{mini}ImageNet (b). For CIFAR100, our proposed UaD-CE outperforms the state of the arts regarding three evaluation indicators. For \textit{mini}ImageNet, though the result of the first session in our proposed framework is inferior to that of \cite{zhu2021self}, our framework outpaces it as novel categories arriving in the following sessions.}
   \vspace{-0.15in}
   \label{fig:cifar+mini}
\end{figure*}

\begin{table*}[t]
\caption{Comparative study on FSCIL and Semi-FSCIL tasks with CUB200. PD is the performance dropping rate from the first session to the last session. Average Acc. is the average performance of all the encountered sessions.}
 \vspace{-0.10in}
\centering
\setlength{\tabcolsep}{11pt}
\resizebox{18.0cm}{!}{
\tabcolsep 0.06in
\begin{tabular}{cccccccccccccccll}
\toprule[1.3pt]
\multirow{2}{*}{\makecell[c]{\textbf{Task}}} & \multirow{2}{*}{\textbf{Method}}& \multicolumn{11}{c}{\textbf{Session ID}} & \multirow{2}{*}{\textbf{PD$\downarrow$}} & \multirow{2}{*}{\makecell[c]{\textbf{Average} \\ \textbf{Acc.}}}\\ 
\cline{3-13} 
 & &  \textbf{1} & \textbf{2} &\textbf{3}
 &\textbf{4} & \textbf{5}& \textbf{6}&\textbf{7} & \textbf{8}& \textbf{9}& \textbf{10} &\textbf{11} 
\\
\midrule[1.3pt]
\multirow{15}{*}{\textbf{FSCIL}}& Ft-CNN~\cite{tao2020few} & 68.68 & 44.81& 32.26 &25.83 &25.62 &25.22& 20.84 &16.77& 18.82 &18.25 & 17.18 & 51.50 & 28.57\\
&Joint-CNN~\cite{tao2020few} & 68.68& 62.43& 57.23& 52.80& 49.50& 46.10& 42.80& 40.10& 38.70& 37.10& 35.60 &  33.08 & 48.28\\
&iCaRL,2017~\cite{rebuffi2017icarl} &68.68& 52.65& 48.61& 44.16& 36.62& 29.52& 27.83& 26.26& 24.01& 23.89& 21.16 & 47.52 & 36.67\\
&EEIL,2018~\cite{castro2018end} &68.68 &53.63 & 47.91 & 44.20& 36.30 & 27.46&25.93& 24.70& 23.95& 24.13& 22.11 &46.57 & 36.27\\
&NCM,2019~\cite{hou2019learning}  &68.68 &57.12 & 44.21 &28.78 &26.71 &25.66 &24.62 &21.52 &20.12 &20.06 &19.87 & 48.81 & 32.49\\
&TOPIC,2020~\cite{tao2020few} &68.68 &62.49 &54.81 &49.99 &45.25 &41.40 &38.35 &35.36&32.22 &28.31 &26.28 & 42.40 & 43.92\\
&SAKD,2021~\cite{cheraghian2021semantic} &68.23& 60.45 &55.70& 50.45& 45.72 &42.90& 40.89& 38.77& 36.51& 34.87& 32.96 & 35.27 & 46.13\\
&SPPR,2021~\cite{zhu2021self} &68.68& 61.85& 57.43& 52.68& 50.19& 46.88& 44.65 &43.07& 40.17& 39.63& 37.33 & 31.35 & 49.32\\
&CEC,2021 \cite{zhang2021few} &75.85 &71.94 &68.50 &63.50 &62.43 &58.27& 57.73 &55.81 &54.83 &53.52 &52.28 &23.57 & 61.33\\
&ERDIL,2021~\cite{dong2021few} &73.52& 71.09& 66.13& 63.25& 59.49& 59.89& 58.64& 57.72& 56.15& 54.75& 52.28 &21.24 &61.17\\
&SFMS,2021~\cite{cheraghian2021synthesized} & 68.78& 59.37& 59.32& 54.96& 52.58 & 49.81 &48.09 &46.32 &44.33& 43.43& 43.23 & 25.55 & 51.84\\
&IDLVQ-C,2021 \cite{chen2021incremental} & \textbf{77.37} &\textbf{74.72}& 70.28& 67.13 & 65.34 &63.52& 62.10 &61.54 &59.04& 58.68& 57.81 & 19.56 & 65.18 \\
& FACT, 2022 \cite{zhou2022forward} & 75.90 & 73.23 & 70.84 & 66.13 & \textbf{65.56} & 62.15 & 61.74 & 59.83 & 58.41 & 57.89 & 56.94 & 18.96 & 64.42 \\
& MetaFSCIL, 2022~\cite{chi2022metafscil} & 75.90 & 72.41 & 68.78 & 64.78 & 62.96 & 59.99 & 58.30 & 56.85 & 54.78 & 53.82 & 52.64 & 23.26 & 61.92\\

& Liu \etal, 2022~\cite{liu2022few} & 75.90 & 72.14 & 68.64 & 63.76 & 62.58 & 59.11 & 57.82 & 55.89 & 54.92 & 53.58 & 52.39 & 23.51 & 61.52 \\

\midrule[1.3pt]
\multirow{7}{*}{\textbf{Semi-FSCIL}}&SS-iCaRL~\cite{cui2021semi} & 69.89 & 61.24 &55.81&50.99&48.18&46.91&43.99&39.78&37.50&34.54& 31.33 & 38.56 & 47.29\\
&SS-NCM~\cite{cui2021semi}&69.89&61.91&55.51&51.71&49.68&46.11&42.19&39.03&37.96&34.05& 32.60 &37.24 & 47.33 \\
&SS-NCM-CNN~\cite{cui2021semi} &69.89&64.87&59.82&55.14&52.48&49.60 &47.87&45.10 &40.47&38.10 &35.25 & 34.64 & 50.78  \\
&Semi-SPPR &68.44& 61.66& 57.11& 53.41& 50.15& 46.68& 44.93 &43.21& 40.61& 39.21& 37.43 & 31.01 & 49.34\\
&Semi-CEC & 75.82& 71.91 & 68.52 & 63.53 & 62.45 & 58.27 & 57.62 & 55.81 & 54.85 & 53.52 & 52.26 & 23.56 & 61.32\\
& Us-KD~\cite{cui2022uncertainty} & 74.69 & 71.71 & 69.04 & 65.08 & 63.60 & 60.96 & 59.06 & 58.68 & 57.01 & 56.41 & 55.54 & 19.15 & 62.89\\
& \textbf{UaD-CE (ours)} & 75.17 & 73.27 & \textbf{70.87} & \textbf{67.14} & 65.49 & \textbf{63.66} & \textbf{62.42} & \textbf{62.55} &\textbf{60.99} &\textbf{60.48} & \textbf{60.72} & \textbf{14.45} & \textbf{65.70}\\
\bottomrule[1.3pt]
\end{tabular}
}
\vspace{-0.035in}
\label{table:cub200}
\end{table*}

\begin{table*}[t]
\caption{Comparative study on the Semi-FSCIL task with CIFAR100 and \textit{mini}ImageNet dataset.} 
\centering
\setlength{\tabcolsep}{11pt}
\resizebox{18.0cm}{!}{
\tabcolsep 0.06in
\begin{tabular}{ccccccccccccccll}
\toprule[1.3pt]
\multirow{2}{*}{\textbf{Dataset}}& \multirow{2}{*}{\textbf{Task}} & \multirow{2}{*}{\textbf{Method}} & \multicolumn{9}{c}{\textbf{Session ID}} & \multirow{2}{*}{\textbf{PD$\downarrow$}} & \multirow{2}{*}{\makecell[c]{\textbf{Average} \\ \textbf{Acc.}}}\\ 
\cline{4-12} 
 & & & \textbf{1} & \textbf{2} &\textbf{3}
 &\textbf{4} & \textbf{5}& \textbf{6}&\textbf{7} & \textbf{8}& \textbf{9} 
\\
\midrule[1.3pt]
\multirow{8}{*}{\rotatebox{90}{\textbf{CIFAR100}}}& \multirow{3}{*}{\textbf{FSCIL}}
& SPPR,2021~\cite{zhu2021self}  & \textbf{76.68} & \textbf{72.69} &	67.61 &	63.52 &	59.18 &	55.82 &	53.08 & 50.89 &	48.12 & 28.56 & 60.84\\
& & CEC,2021 \cite{zhang2021few}  &73.03 & 70.86 & 65.20 & 61.27 & 58.03 & 55.53 & 53.17 & 51.19 & 49.06 & 23.97 & 59.70\\
& & UaD-CE (ours) & 75.55 & 71.78 & 65.47 & 62.83& 55.56 & 55.08 & 50.11 & 46.35 & 40.46 & 35.09 & 58.13 \\
\cline{2-14}
&\multirow{6}{*}{\textbf{Semi-FSCIL}} & Semi-SPPR  &76.68& 72.63& 67.59& 63.69 & 59.24 & 56.02 & 53.23 & 50.46 & 48.29 & 28.39 & 60.87\\
& & Semi-CEC & 73.03 & 70.72 & 65.79 & 61.91  & 58.64 & 55.84 & 53.70 & 51.37 & 49.37& 23.66 & 60.04\\
& & SS-iCaRL~\cite{cui2021semi}&64.13 & 56.02 & 51.16 & 50.93& 43.46 & 41.69 & 38.41 & 39.25 &  34.80 & 29.33 & 46.65\\
& & SS-NCM-CNN~\cite{cui2021semi} &64.13 & 62.29&  61.31& 57.96& 54.26&  50.95&49.02& 45.85&44.59& \textbf{19.54} & 54.51\\
& & Us-KD~\cite{cui2022uncertainty} & 76.85 & 69.87 & 65.46 & 62.36 & 59.86 & 57.29 & 55.22 & 54.91 & 54.42 & 22.43 & 61.80 \\
& & UaD-CE (ours) & 75.55 & 72.17 & \textbf{68.57} & \textbf{65.35} & \textbf{62.80} & \textbf{60.27} & \textbf{59.12} & \textbf{57.05} & \textbf{54.50}  & 21.05 & \textbf{63.93}\\
\midrule[1.3pt]
\multirow{8}{*}{\rotatebox{90}{\textbf{\textit{mini}ImageNet}}}& \multirow{3}{*}{\textbf{FSCIL}}& SPPR,2021~\cite{zhu2021self}  &  \textbf{80.27} & \textbf{74.22} & 68.89 & 64.43 & \textbf{60.54} & 56.82 &	\textbf{53.81} & 51.22 &	48.54 &  31.73 & 62.08\\
&& CEC,2021 \cite{zhang2021few}  &72.22 &67.06 &63.17 &59.79 &56.96 &53.91& 51.36 &49.32 &47.60 & 24.62 & 57.93\\
& & UaD-CE (ours) & 72.35 & 66.83 & 61.94 & 58.48& 55.77 & 52.20 & 49.96 & 47.96 & 46.81 & 25.54 & 56.92\\
\cline{2-14} 
& \multirow{6}{*}{\textbf{Semi-FSCIL}} & Semi-SPPR &  80.10&74.21&\textbf{69.31}&\textbf{64.83}& 60.53 &\textbf{57.36}&53.70&\textbf{52.01}&49.61&
30.49 & \textbf{62.41} \\
& & Semi-CEC,2021 & 71.91 & 66.81 & 63.87 & 59.41  & 56.42 & 53.83 & 51.92 & 49.57 & 47.58 & 24.33 & 57.92 \\
& & SS-iCaRL~\cite{cui2021semi} & 62.98& 51.64& 47.43& 43.92& 41.69& 38.74& 36.67& 34.54& 33.92& 29.06 & 43.50 \\
& & SS-NCM-CNN~\cite{cui2021semi} & 62.98& 60.88& 57.63& 52.8& 50.66& 48.28& 45.27& 41.65& 40.51 & 22.47 &51.26 \\
& & Us-KD~\cite{cui2022uncertainty}  & 72.35 & 67.22 & 62.41 & 59.85 & 57.81 & 55.52 & 52.64 & 50.86 & 50.47 & 21.88 & 58.79\\
& & UaD-CE (ours) & 72.35 & 66.91 & 62.13 & 59.89 & 57.41 & 55.52 & 53.26 & 51.46 &\textbf{50.52}  & \textbf{21.83} & 58.82\\
\bottomrule[1.3pt]
\end{tabular}
}
\vspace{-0.1in}
\label{table:semi}
\end{table*}

\begin{table*}[t]
\caption{Prediction accuracy of base and novel classes on three benchmark datasets.}
\vspace{-0.1in}
\centering
\setlength{\tabcolsep}{11pt}
\resizebox{18.0cm}{!}{
\tabcolsep 0.06in
\begin{tabular}{cccccccccccccccc}
\toprule[1.3pt]
\multirow{2}{*}{\textbf{Dataset}}& \multirow{2}{*}{\textbf{Task}} & \multirow{2}{*}{\textbf{Method}} & \multirow{2}{*}{\makecell[c]{\textbf{Classes}}}& \multicolumn{9}{c}{\textbf{Session ID}}\\ 
\cline{5-15} 
&  &  & & \textbf{1} & \textbf{2} &\textbf{3}
 &\textbf{4} & \textbf{5}& \textbf{6}&\textbf{7} & \textbf{8}& \textbf{9} &\textbf{10}& \textbf{11}
\\
\midrule[1.3pt]
\multirow{15}{*}{\rotatebox{90}{\textbf{CIFAR100}}}& \multirow{6}{*}{\textbf{FSCIL}} & \multirow{3}{*}{SPPR,2021~\cite{zhu2021self}}& Base  & 76.68 & 77.48
& 77.32 & 76.95	&60.25 & 76.42 &75.95 & 75.83 &  75.30 &-&-\\
& & &Novel & - & 1.52
& 9.40	 & 9.80	& 7.45 & 7.52 & 7.33 &  8.14 & 7.35&-&- \\
& & & All & \textbf{76.68} & \textbf{72.69} &	67.61 &	63.52 &	59.18 &	55.82 &	53.08 & 50.89 &	48.12 &-&-\\
\cline{3-15}
& & \multirow{3}{*}{CEC,2021 \cite{zhang2021few}}  
&Base & 73.03 & 73.77
 & 71.52	 &70.78	& 70.07&69.28	 &68.70 	 & 68.45 &67.78&-&-\\
& & &Novel & - & 36.00
& 27.30	 & 22.67 & 21.90 & 22.52 & 22.10	 & 21.60 &  20.98&-&- \\
& & &All&73.03 &70.86 &65.20 &61.27 &58.03 &55.53& 53.17 &51.19 &49.06&-&- \\
\cline{2-15}
&\multirow{9}{*}{\textbf{Semi-FSCIL}} &  \multirow{3}{*}{SS-iCaRL~\cite{cui2021semi}} & Base & 64.13 & 56.97 & 53.77 & 55.30 & 48.07 & 47.48 & 44.18 & 45.84 & 40.97 &  - & -\\ 
& & & Novel & -& 44.57 & 35.48 & 33.47 & 27.40 & 27.39 & 25.71 & 27.97 & 26.10 & - & - \\
& & & All & 64.13 & 56.02 & 51.16 & 50.93 & 43.46 & 41.69 & 38.41 & 39.25 & 34.80 & - & -\\
\cline{3-15}
& & \multirow{3}{*}{SS-NCM-CNN~\cite{cui2021semi}} & Base & 64.13 & 63.34 & 64.43 & 62.93 & 60.01 & 58.02 & 56.39 & 53.55 & 52.49\\ 
& & & Novel & - & 49.56 & 42.51&38.08&34.21& 33.99&32.81&32.67&33.44&- & -\\
& & & All &64.13 & 62.29&61.31&57.96&54.26&  50.95&49.02&45.85&44.59 & - & -\\
\cline{3-15}
& & \multirow{3}{*}{UaD-CE (ours)} & Base & 75.55 
& 73.20	 & 70.06 & 68.94 &	67.63 & 66.82	 & 65.81 &  65.25 & 61.99 & - & -\\
& & &Novel & - &  \textbf{59.80} & \textbf{59.21} & \textbf{53.00}	 & \textbf{48.30}	& \textbf{46.60} & \textbf{45.73}	 & \textbf{43.00} &  \textbf{42.27} &-&-\\
& & &All & 75.55 & 72.17 & \textbf{68.57} & \textbf{65.35} & \textbf{62.80} & \textbf{60.27} & \textbf{59.12} & \textbf{57.05} &\textbf{54.50}&-&-\\
\bottomrule[1.3pt]
\multirow{15}{*}{\textbf{\rotatebox{90}{\textit{mini}ImageNet}}}& \multirow{6}{*}{\textbf{FSCIL}} & \multirow{3}{*}{SPPR,2021~\cite{zhu2021self}}  
&Base & 80.27& 80.22 & 80.28 & 80.20 & 80.08 &	79.77 &	79.98 &	79.58 &	79.87&-&-\\
& & &Novel & - &2.20  & 0.50
& 1.33	 &1.90	& 1.76&1.47	 &	2.60 & 1.55&-&-   \\
& & & All & \textbf{80.27} & \textbf{74.22} &\textbf{ 68.89} & \textbf{64.43} & \textbf{60.54} &	\textbf{56.82} &	\textbf{53.81} &	51.22 &	48.54&-&-\\
\cline{3-15}
& & \multirow{3}{*}{CEC,2021 \cite{zhang2021few}} & Base  & 72.22 &70.92 & 70.17
 &69.65	& 69.32&68.98 	 &	68.68 & 68.25 &67.87 &-&-\\
& & &Novel &-  & 20.80 &21.20 
& 20.33	 &19.90	& 17.72&16.70	 &	16.86 & 17.20  &-&-\\
& & & All & 72.22 & 67.06 & 63.17 & 59.79 &56.96 &53.91& 51.36 &49.32 &47.60 &-&-\\
\cline{2-15}
&\multirow{9}{*}{\textbf{Semi-FSCIL}} &  \multirow{3}{*}{SS-iCaRL~\cite{cui2021semi}} & Base & 62.98 & 53.63 & 49.85 & 47.69 & 46.11 & 44.10 & 42.18 & 40.34 & 39.93 & - & - \\ 
& & & Novel & - & 27.66 & 32.89 & 28.86 & 26.28 & 25.83 & 24.54 & 24.61 & 25.44 &-& -\\
& & & All & 62.98& 51.64& 47.43& 43.92& 41.69& 38.74& 36.67& 34.54& 33.92 & - & - \\
\cline{3-15}
& & \multirow{3}{*}{SS-NCM-CNN~\cite{cui2021semi}} & Base & 62.98 & 63.23 & 60.57 & 57.33 &  56.03 & 54.98 & 52.08 & 48.64 & 47.69 & - & -\\ 
& & & Novel & - & \textbf{32.61} & \textbf{39.96} & 34.69 & 31.94 & 32.21 & 30.30 & 29.68 & 30.38 & - & -\\
& & & All & 62.98& 60.88& 57.63& 52.80& 50.66& 48.28& 45.27& 41.65& 40.51 & - & -\\
\cline{3-15}
& & \multirow{3}{*}{UaD-CE (ours)} & Base  & 72.35 &  70.27&	 67.36 & 66.06	 & 64.26	& 63.14 & 61.78& 60.71& 59.20&-&-\\
& & &Novel & - & 27.03 & 30.73 & \textbf{35.20}	 & \textbf{36.85}	& \textbf{37.24} & \textbf{36.23}	 &	\textbf{35.60} & \textbf{36.70}  &-&-\\
& & &All & \textbf{72.35} & 66.91 & 62.13 & 59.89 & 57.41 & 55.52 & 53.26 & \textbf{51.46} &\textbf{50.52}  &-&-\\
\bottomrule[1.3pt]
\multirow{15}{*}{\rotatebox{90}{\textbf{CUB200}}} & \multirow{6}{*}{\textbf{FSCIL}} & \multirow{3}{*}{SPPR,2021~\cite{zhu2021self}}  
&Base & 68.16 & 60.67 & 59.87 & 60.13 & 57.82 & 55.85& 53.77&	53.57 & 52.95& 52.43 & 51.73 \\
& & &Novel  & -  & \textbf{60.39} & 45.55 & 30.24 & 31.07 & 27.75 & 26.67& 27.51	 & 25.16 & 24.76&  23.88 \\
& &&All& 68.16& 60.32& 57.11&52.79&49.68&45.95&43.19&42.39& 40.17& 38.93&37.33\\
\cline{3-15}
& & \multirow{3}{*}{CEC,2021 \cite{zhang2021few}} & Base   & 75.82& 74.34& 73.94 & 73.59 & 72.73 & 72.39 & 71.90& 71.28	 & 71.14	 & 70.90 &  70.66\\
& & &Novel  & - & 46.61 & 41.74 & 33.34 & 37.46 & 32.77 &	34.78 &	35.06 & 33.23 & 34.98 & 34.17\\
& & &All & \textbf{75.82} & 71.91& 68.52&63.53& 62.45& 58.27 & 57.62& 55.81& 54.85&53.52&52.26 \\
\cline{2-15}
&\multirow{9}{*}{\textbf{Semi-FSCIL}} &  \multirow{3}{*}{SS-iCaRL~\cite{cui2021semi}} & Base & 69.89 & 62.32 &  60.62 & 58.99 & 58.59 & 57.77 & 59.88 & 56.21 & 54.46 &50.54 & 46.11\\ 
& & & Novel & - & 53.22 & 32.38 & 24.07 & 22.76 & 23.34 & 17.58 & 16.40 & 16.39 & 16.13 & 16.32\\
& & & All & 69.89 & 61.24 &55.81&50.99&48.18&46.91&43.99&39.78&37.50&34.54& 31.33 \\
\cline{3-15}
& & \multirow{3}{*}{SS-NCM-CNN~\cite{cui2021semi}} & Base & 69.89 & 65.80 & 64.97 & 63.79&63.81&61.08& 65.24&63.73&58.77&55.74&51.88\\ 
& & & Novel & - & 56.37 & 34.70 & 26.03 & 24.04 & 24.68 & 19.14 & 18.60 & 17.70 & 17.79 & 18.36\\
& & & All & 69.89 & 64.87 & 59.82 & 55.14 & 52.48 & 49.60 & 47.87 & 45.10 & 40.47 & 38.10 & 35.25  \\
\cline{3-15}
& & \multirow{3}{*}{UaD-CE (ours)} & Base &75.17  &74.58 & 73.78	 &	72.97 & 71.33  & 70.88 &69.76	 &  69.48& 68.26 & 68.82&68.47 \\
& & &Novel & - & 59.86 & \textbf{56.18} & \textbf{47.80}	 &	\textbf{51.12} &\textbf{49.52 } &\textbf{50.40}	 &	\textbf{52.86} & \textbf{52.09} & \textbf{51.14} & \textbf{53.14} \\
& & &All& 75.17 & \textbf{73.27} & \textbf{70.87} & \textbf{67.14} & \textbf{65.49} & \textbf{63.66} & \textbf{62.42} & \textbf{62.55} &\textbf{60.99} &\textbf{60.48} & \textbf{60.72} \\
\bottomrule[1.3pt]
\end{tabular}
}
\vspace{-0.10in}
\label{table:acc}
\end{table*}

\subsection{Construction Details of the Comparative Study on Semi-FSCIL Task.} For the fair comparison, we also implemented Semi-FSCIL setting based on existing methods: SPPR \cite{zhu2021self}, and CEC \cite{zhang2021few}, which is regarded as Semi-SPPR and Semi-CEC. For SPPR and CEC, the backbone is frozen after the first session, then the average feature mean (\ie, prototypes) of a particular novel category is computed by obtaining features with this backbone in the following sessions. In the first session, base classes are also represented by prototypes. To further build a unified classifier, prototypes of all categories encountered so far are fed into a graph network \cite{zhang2021few} or an attention module \cite{zhu2021self}. When incorporating unlabeled data, we first compute prototypes based on limited labeled samples, then update them with unlabeled data according to the distance between the current prototypes and features of unlabeled data.

\subsection{Comparative Studies}

The comparative experimental results encompass three perspectives: (1) To certify that our proposed UaD-CE for Semi-FSCIL can benefit from harnessing unlabeled samples, we compared UaD-CE with the state of the arts for FSCIL setting; (2) For the fairness issue, we compared with the exist Semi-FSCIL method~\cite{cui2021semi} and implemented Semi-FSCIL setting with existing new FSCIL methods to elaborate that our proposed framework can gain more from unlabeled samples; (3) To illustrate that our UaD-CE possesses the superior performance on mitigating overfitting issue and the bias to base categories, we present the accuracy of base and novel categories respectively, and compare them with the state of arts. It is worth noting that the evaluation process for each dataset was repeated 5 times, and we reported the average overall accuracy.

\vspace{0.1in}
\noindent \textbf{Comparative study on the FSCIL task.} The results of comparison FSCIL methods are directly quoted from original papers to facilitate fair comparison. CIFAR100 and \textit{mini}ImageNet results are presented in Figure~\ref{fig:cifar+mini} (a) and Figure~\ref{fig:cifar+mini} (b), respectively. Besides, Table~\ref{table:cub200} presents the results of CUB200. When comparing UaD-CE with the state of the arts on CIFAR100, \textit{mini}ImageNet and CUB200 for FSCIL setting, we conducted experiments with regard to three evaluation indicators: (1) The final overall accuracy ($\%$). Our proposed framework exceeds all other methods on three datasets and even outperforms the ``Joint-CNN'' methods to a large extent, which is defined as the upper bound in \cite{tao2020few}. Especially, our proposed UaD-CE achieves 54.50\% on CIFAR100, 50.52\% on \textit{mini}ImageNet and 60.72\% on CUB200 in the final session, surpassing the state of the arts by around 4\% and 3\%. (2) For the performance dropping rate (PD), the proposed UaD-CE also gains remarkable capability. Although for the first session, our method cannot exceed the state of the arts on \textit{mini}ImageNet and CUB200, the accuracies obtained by our framework go through slight descending curves, which indicates that old knowledge can be better preserved in our proposed UaD-CE. (3) The average accuracy ($\%$) of all the sessions. The UaD-CE obtains advanced results on CIFAR100 (63.93\%) and CUB200 (65.70\%). As for \textit{mini}ImageNet, the best performance of average accuracy is shown in \cite{zhu2021self}, which mainly results from the higher accuracy of the first session. As CIL aims to mitigate catastrophic forgetting on old categories, the attenuation of accuracy (\ie, PD) is more convincing than the average accuracy for evaluating CIL methods. Though the result of the first session in our proposed framework is inferior to that in \cite{zhu2021self}, our UaD-CE outpaces it as novel categories arriving in the following sessions.

\noindent \textbf{Comparative study on the Semi-FSCIL task.} To explain that unlabeled samples can contribute more to FSCIL with our proposed framework, we present the comparative results with the Semi-FSCIL method~\cite{cui2021semi}. Furthermore, the same unlabeled instances were fairly incorporated into existing FSCIL methods, referred to Semi-FSCIL. CEC~\cite{zhang2021few} and SPPR \cite{zhu2021self} are advanced works for FSCIL, and they also release codes \footnote{https://github.com/icoz69/CEC-CVPR2021}\footnote{https://github.com/zhukaii/SPPR}. These codes can reduce deviations brought by implementation details from original papers and be reliant on conducting Semi-CEC and Semi-SPPR. Results of Semi-CEC and Semi-SPPR are presented on Table~\ref{table:cub200} for CUB200, and Table~\ref{table:semi} for CIFAR100 and \textit{mini}ImageNet. In view of the results, the performance of FSCIL boosts to a greater extent by harnessing unlabeled samples with our UaD-CE framework. However, when the same number of unlabeled samples are combined to Semi-CEC and Semi-SPPR, a slight enhancement is obtained. The reason is that CEC and SPPR heavily rely on refining the acquired latent features by the graph structure \cite{zhang2021few} or attention modules \cite{zhu2021self}. This particularity undermines the role of the latent representative capability, where the semi-supervised learning technique fundamentally contributes. Compared with Us-KD~\cite{cui2022uncertainty}, UaD-CE outperforms it regarding three evaluation indicators, which illustrates that considering the uncertainty issue during distillation is more applicable and efficient than that of in semi-supervised learning process.

\vspace{0.1in}
\noindent \textbf{Comparative results of base and novel categories.} 
To reveal that UaD-CE is efficacious to conquer the overfitting challenges and mitigate the bias to base categories, we separately exhibit the accuracy of base categories (\ie, $\mathcal{C}_1$) and novel categories (\ie, $\mathcal{C}_2 \cup, ..., \mathcal{C}_n$) in Table~\ref{table:acc}. For CEC and SPPR, we use
the released codes to conduct comparison experiments on base and novel categories. Models can better remember the performance of base categories with the large-scale dataset, while the overfitting problem is still severe among novel classes with limited labeled samples. Although the overall performance is roughly satisfactory, these methods can not resolve two issues in the FSCIL task at the same time, which bring out the severe classification bias to base categories. Surprisingly, our proposed UaD-CE can mitigate overfitting on novel categories while maintaining the performance of the base ones.

\begin{figure}[t]
  \centering
   \includegraphics[width=1.0\linewidth]{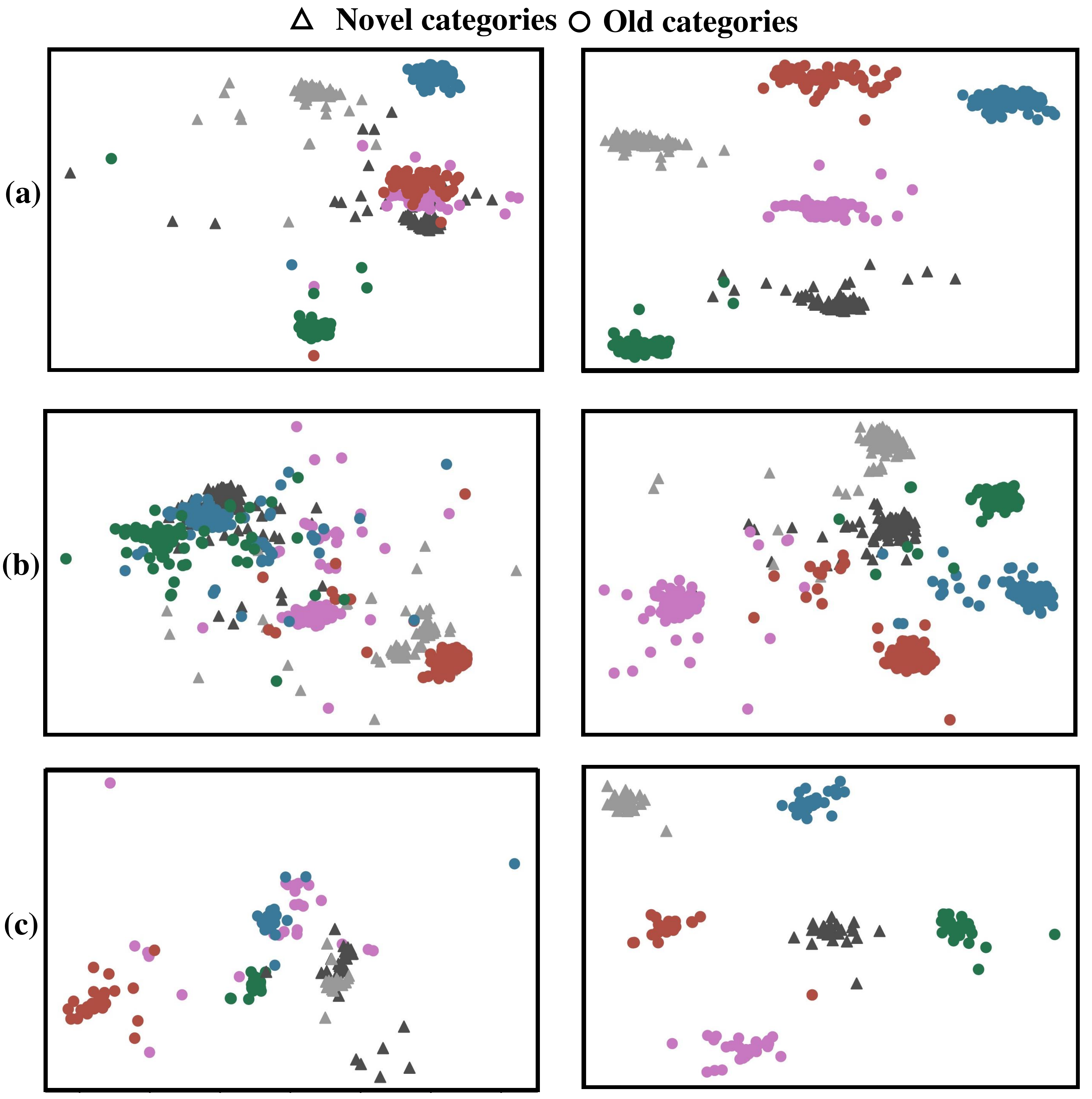}
   \vspace{-0.15in}
   \caption{(a)-(c) The t-SNE visualization of the features for CIFAR100 (a), \textit{mini}ImageNet (b), and CUB200 (c).}
   \vspace{-0.20in}
   \label{fig:remark}
\end{figure}

\begin{figure}[t]
  \centering
   \includegraphics[width=0.8\linewidth]{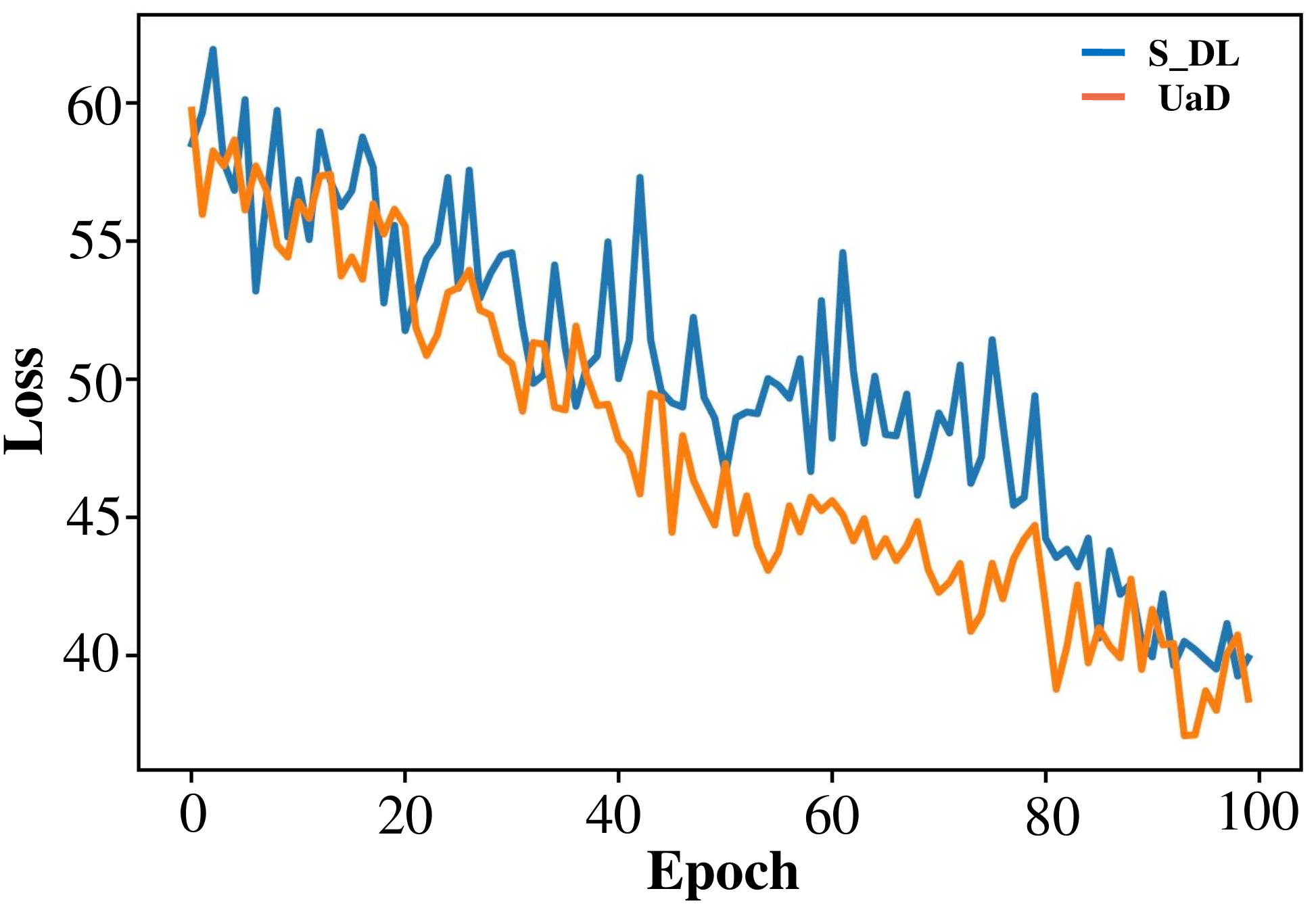}
   \vspace{-0.10in}
   \caption{\textcolor{black}{Loss curves of the second session in CIFAR100. \textit{S\_DL} means standard knowledge distillation.}}
   \vspace{-0.20in}
   \label{fig:loss}
\end{figure}

\noindent \textbf{Further remark}. We present the visualized t-SNE \cite{Van2008visual} results in Figure~\ref{fig:remark} (a)-(c). The t-SNE results of our methods (the right) have clearer decision boundaries of the old and new categories than that of iCaRL~\cite{rebuffi2017icarl} (the left) used the standard distillation technique (\ie, Equation \ref{eq:loss_total_old}). In addition, we visualize the loss of the training on CIFAR100 to illustrate the superiority of the UaD module in Figure~\ref{fig:loss}. From the loss curves, it can be shown that curve (b) representing UaD experiences relatively smaller fluctuations. This fact indicates that the gradients generated by computing distillation losses on the uncertainty-guided refined exemplar set have similar decline directions, \ie, in most cases, valid old knowledge from the reference model is transferred to the target model.

\begin{figure*}[t]
  \centering
 \includegraphics[width=1.0\linewidth]{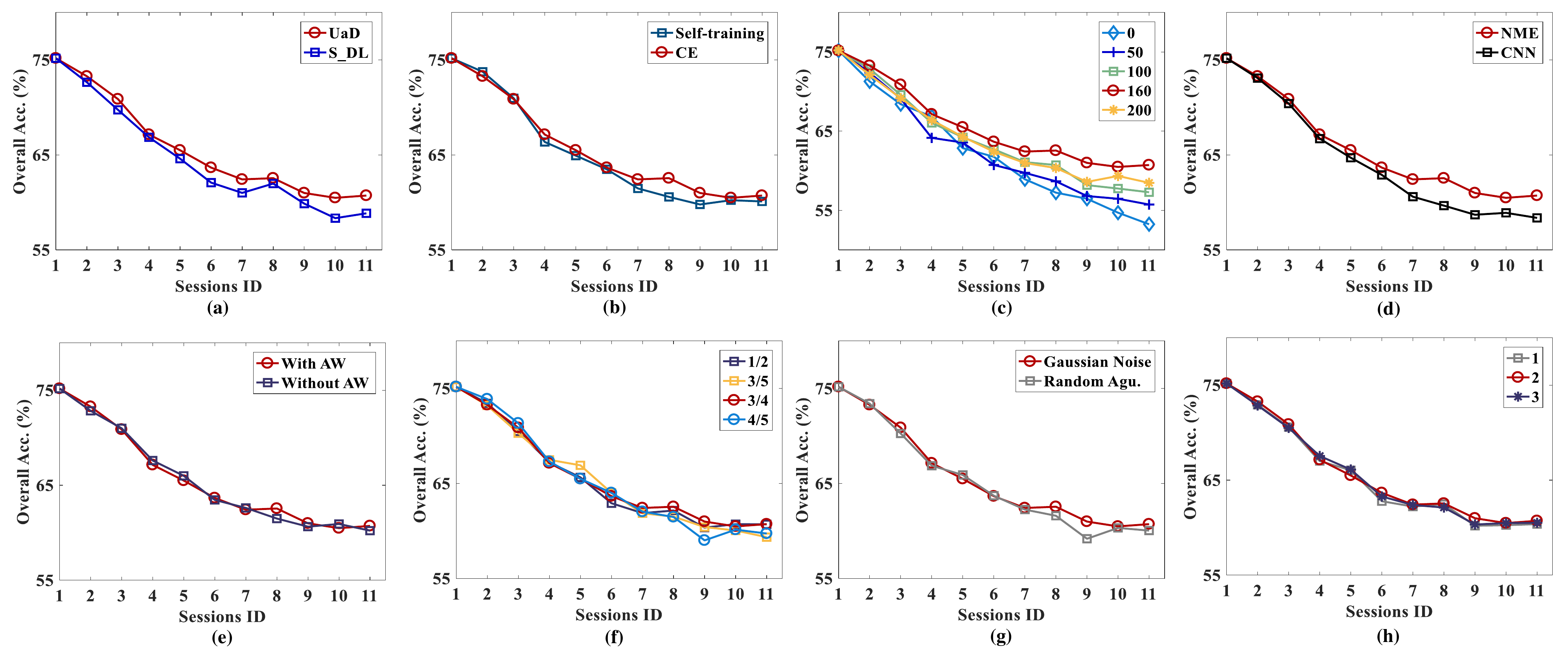}
   \vspace{-0.3in}
   \caption{Ablation study on modules and hyperparameters in UaD-CE with CUB200.}
   \vspace{-0.1in}
   \label{fig:ablation}
\end{figure*}

\begin{table*}[t]
\caption{Ablation study results on the number of unlabeled samples incorporated in each incremental learning session. \textit{\#Unlabeled} represents the number of unlabeled samples incorporated into each incremental session. \textit{Prop (\%)} refers to the proportion of the number of labeled samples to the number of all samples.}
\vspace{-0.1in}
\centering
\setlength{\tabcolsep}{11pt}
\resizebox{18.0cm}{!}{
\tabcolsep 0.06in
\begin{tabular}{ccccccccccccccccc}
\toprule[1.3pt]
\multirow{2}{*}{\textbf{Dataset}}& \multirow{2}{*}{\textit{\textbf{\#Unlabeled}}} & \multirow{2}{*}{\makecell[c]{\textit{\textbf{Prop (\%)}}}}& \multicolumn{9}{c}{\textbf{Session ID}}\\ 
\cline{4-14} 
&  &  & \textbf{1} & \textbf{2} &\textbf{3}
 &\textbf{4} & \textbf{5}& \textbf{6}&\textbf{7} & \textbf{8}& \textbf{9} &\textbf{10}& \textbf{11}
\\
\midrule[1.3pt]
\multirow{3}{*}{\textbf{CIFAR100}} & 250 & 9.10  & 75.55& 71.49&67.76&64.00&61.02&58.35&57.23&55.46&53.65 &-&-\\
& \textbf{350} & \textbf{6.70} & 75.55 & \textbf{72.17} & \textbf{68.57} & \textbf{65.35} & \textbf{62.80} & \textbf{60.27} & \textbf{59.12} & \textbf{57.05} &\textbf{54.50}&-&- \\
& 450 & 5.30 & 75.55&71.67&67.67&64.20&60.70&58.51&56.78&55.27&53.67 &-&-\\

\bottomrule[1.3pt]
\multirow{3}{*}{\textbf{\textit{mini}ImageNet}}   
& 100 & 20.00 & 72.35& 61.89&58.83&56.39&53.14&51.01&49.56&48.62&-&- \\
 & \textbf{160} & \textbf{13.51} &\textbf{72.35} & \textbf{66.91} & \textbf{62.13} & \textbf{59.89} & \textbf{57.41} & \textbf{55.52} & \textbf{53.26} & \textbf{51.46} &\textbf{50.52}&-&-   \\
& 200 & 11.11 & 72.35&66.62&61.86&58.76&56.27&53.81&51.83&50.83&49.78&-&-\\

\bottomrule[1.3pt]
\multirow{3}{*}{\textbf{CUB200}}  
& 100  & 33.33 &75.17&72.86& 69.59&66.04&64.23&62.70&61.08&60.72&58.18& 57.75& 57.28 \\
& \textbf{160} & \textbf{23.81}  &  75.17 & \textbf{73.27} & \textbf{70.87} & \textbf{67.14} & \textbf{65.49} & \textbf{63.66} & \textbf{62.42} & \textbf{62.55} &\textbf{60.99} &\textbf{60.48} & \textbf{60.72}\\
& 200 & 20.00 & 75.21&72.10&69.15&66.40&64.28&62.40&60.97&60.34&58.59&59.34&58.47\\

\bottomrule[1.3pt]
\end{tabular}
}
\vspace{-0.10in}
\label{table:ablation_num_unlabeled}
\end{table*}

\vspace{-0.15in}
\subsection{Ablation Studies}

To demonstrate the efficacy of UaD-CE framework, we conducted extensive ablation experiments on the proposed modules and hyperparameters.

\noindent \textbf{Efficacy of UaD module.}
The UaD-CE framework in this paper is set up based on the knowledge distillation technique for the class-incremental learning purpose, and the proposed UaD module is proposed to make this technique more applicable to Semi-FSCIL. The common form of knowledge distillation usually follows the distillation loss defined in Equation~\ref{eq:loss_total_old} ~\cite{hinton2015distilling}, and we term it standard knowledge distillation (\textit{S\_DL}). We evaluated our framework by replacing UaD module with \textit{S\_DL}, and the comparative result is illustrated in Figure~\ref{fig:ablation} (a). With the UaD module, the overall classification accuracy exceeds that of the standard knowledge distillation in all incremental sessions, proving the prominent adaptability of the UaD module to Semi-FSCIL. This adaptability is achieved due to the quality and quantity of old knowledge we consider in the distillation process. With the UaD module, the negative effect brought by unlabeled data can be mitigated to some extent, \ie, unlabeled data can better assist the knowledge distillation process in memorizing the old knowledge.

\noindent \textbf{Efficacy of CE module.} As for effectively harnessing unlabeled data in each incremental session to alleviate the overfitting issue, we propose the CE module in which the main component is class-balance self-training.  
We analyzed the function of class-balanced operation in CE module by comparing the performance with traditional self-training~\cite{grandvalet2004semi}, which is presented in Figure~\ref{fig:ablation} (b). The curve with CE module experiences a slight decline. It demonstrates that the CE module can enhance the overall classification performance by taking the model's learning status into consideration. In each incremental session, the difficulty of learning varies for different classes, and it also refers to the learning status of the model. Therefore, in our proposed CE module, when selecting unlabeled data based on the predictions, we set a specific threshold for each class, which can mitigate the performance imbalance. The difficulty of learning also varies among different incremental sessions. This is why the curve experiences a slight increase in the eighth session.

\noindent \textbf{Impact of number of incorporated unlabeled samples.} Since we discuss the adaptability issue of semi-supervised learning to the FSCIL task, the extreme value of incorporated unlabeled data is explored in our experiments so that the performance can achieve the greatest improvement. The ablation results on the number of incorporated unlabeled samples are exhibited in Figure~\ref{fig:ablation} (c). With different numbers of unlabeled samples, the overall classification accuracy can increase to varying degrees when compared with the performance obtained without extra unlabeled data. In this way, it can be summarized that the performance of FSCIL can profit from the combination of unlabeled data. With 160 unlabeled samples, the proposed UaD-CE obtains the surpassing performance. There are mainly two reasons why more unlabeled data is not better: (1) Unlabeled data may bring the noise to the model training due to its quality and the uncertainty of obtaining its pseudo label. Consequently, the trade-off between the profit and the noise brought by unlabeled data exists in Semi-FSCIL tasks. In our future work, we will seek for advanced technique to mitigate the negative effects brought by unlabeled data in Semi-FSCIL. (2) For distillation-based incremental learning methods, the trade-off between the performance of old and new classes should be taken into consideration. In particular, Tao~\etal~\cite{tao2020few} point out that this trade-off is more challenging for FSCIL because the large learning rate was required to learn novel classes with limited labeled data. Though unlabeled data is combined into each incremental learning session to relax this requirement, we still need to handle the trade-off issue carefully. Moreover, we also illustrate the ablation study results in Table~\ref{table:ablation_num_unlabeled} with the numerical representation and the proportion (\%) of labeled samples. The best proportion varies for different datasets, and it may result from the nature of the dataset itself and the incremental pattern. We will continue exploring the root causes affecting this proportion in future work.

\vspace{0.05in}
\noindent \textbf{Impact of classification head.}
In the default setting of UaD-CE framework, we apply the nearest-mean-of-exemplars (NME) classification. In this ablation, we consider the CNN predictions and give the comparative result in Figure~\ref{fig:ablation} (d). 
With increasing novel categories with limited labeled instances, the NME classification outperforms the CNN prediction significantly from the seventh session. NME classification is to measure the distance between the features of test images and prototypes (\ie, the class mean in the feature level), where the computation is only related to parameters of the backbone. However, the CNN prediction also requires putting the features into the fully-connected layer, and the prediction is then conducted based on the final outputs. Aiming at the incremental learning pattern, the fully-connected layer has to expand to fit the increase of total classes. For a specific session, the parameters of the backbone are optimized in all encountered sessions, while the parameters related to the new neurons of the fully-connected layers are only updated in the current session, which can explain why the NME classification achieves the superior performance to the CNN prediction.

\vspace{0.05in}
\noindent \textbf{Impact of $\bm\zeta$ in UaD module.}
One of the challenges for the distillation-based framework in FSCIL is the trade-off between old and new classes. With the number of seen classes increasing, the model should memorize more old knowledge for the overall classification performance. The weight $\zeta$ tradeoffs the contribution of previous knowledge to the current task for enabling the overall classification ability. Figure~\ref{fig:ablation} (e) illustrates the performance without this dynamically changeable weight (Without AW). We can conclude that the adaptive weight for distillation loss is effective for Semi-FSCIL. 

\vspace{0.05in}
\noindent \textbf{Impact of $\bm\tau$ in UaD module.}
Hyperparameter $\tau$ is the threshold of the uncertainty, and it determines how many reliable exemplars of previous classes are reserved totally in a specific following session for the knowledge distillation. This value also determines the quantity of old knowledge considered when we conduct the adaptive weights. We increase this number from 0.5 to 0.8, and report the results in Figure~\ref{fig:ablation} (f). The superior overall classification performance is achieved when we remain the first three-quarter of exemplars based on the uncertainty. 

\vspace{0.05in}
\noindent \textbf{Study of noise in uncertainty estimation part.} We apply the test-time augmentation method for the uncertainty estimation, and Gaussian noise is added to the original images for the augmentation. Besides the Gaussian noise used in the uncertainty estimation part, we conduct the ablation experiments with random augmentation, \ie, random crop and random flip. As shown in Figure~\ref{fig:ablation} (g), using Gaussian noise for uncertainty estimation is more valid in UaD-CE. Since the model is trained with a large-scale dataset, the model can have a relatively robust discriminative capability. In this way, random augmentation is too weak to generate the diverse predictions of a specific image, thus making the uncertainty-aware module less salient to the model performance. 

\vspace{0.05in}
\noindent \textbf{Impact of ${\bm\zeta}^{\bm{base}}$ in UaD module.} The ablation study result on $\zeta^{base}$ is given in Figure~\ref{fig:ablation} (h). Hou~\etal~\cite{hou2019learning} also apply the adaptive weight of the distillation loss and point out that $\zeta^{base}$ is constant for a specific dataset. We also follow this configuration in our experiment. However, in UaD-CE framework, $\zeta^{base}$ is not sensitive to this parameter, which demonstrates that the last two components in $\zeta$ are more important than $\zeta^{base}$. In future work, we will explore better forms of adaptive weight.

\begin{figure}[t]
  \centering

   \includegraphics[width=0.8\linewidth]{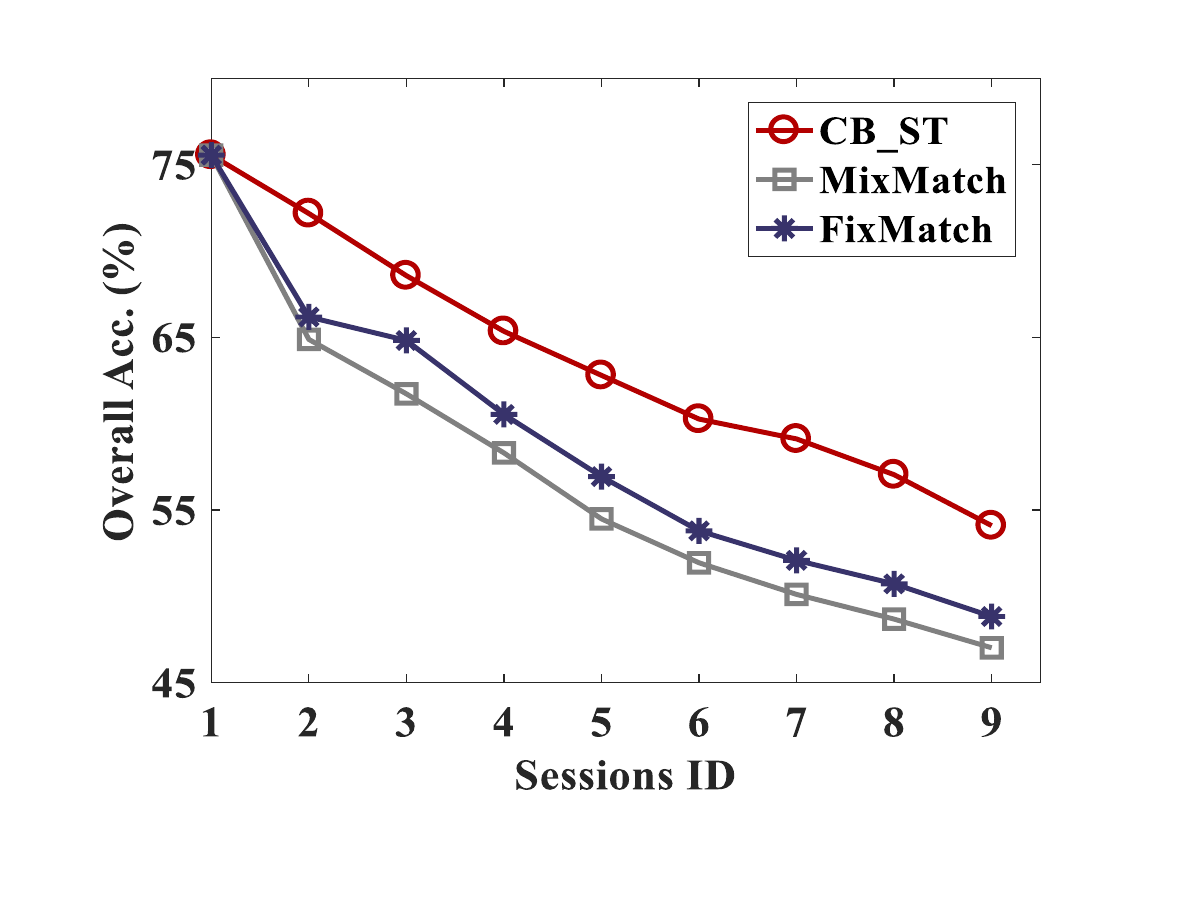}
   \vspace{-0.15in}
   \caption{Ablation results of CIFAR100 on semi-supervised learning method.}
   \vspace{-0.20in}
   \label{fig:ablation-semi}
\end{figure}

\vspace{0.05in}
\noindent \textbf{Comparison with more semi-supervised learning methods.}
In the UaD-CE framework, we apply
class-balanced self-training (CB\_ST) when harnessing unlabeled data in a semi-supervised manner. To illustrate the efficacy of class-balanced self-training, we replace it with the consistency regularization-based methods, \ie, MixMatch~\cite{berthelot2019mixmatch} and FixMatch~\cite{sohn2020fixmatch}. These ablation study results are presented in Figure~\ref{fig:ablation-semi}. It can be seen that the performance obtained with MixMatch and FixMatch is inferior to that of class-balanced self-training. Since our proposed framework is knowledge distillation-based, the performance trade-off between
old and new classes in FSCIL is hard to handle, which is pointed out in~\cite{tao2020few}. For consistency regularization-based methods, the consistency loss is generated on unlabeled samples. When this loss is combined with distillation loss, it may generate gradients for learning new classes in different directions from that of distillation loss, which is not conducive to preserving the old knowledge. We will seek the solutions that can make advanced consistency regularization-based semi-supervised learning methods efficacious in Semi-FSCIL.

\vspace{-0.05in}
\section{Conclusion and Future Work}

In this paper, we focus on tackling the unresolved adaptability issue of semi-supervised learning to the FSCIL task and propose a simple yet efficient Semi-FSCIL framework named UaD-CE, which includes CE and UaD modules. CE is presented to address the overfitting and bias issues in FSCIL by incorporating unlabeled samples with class-balanced self-training. Furthermore, to efficiently address the catastrophic forgetting problem, we conduct the UaD module with uncertainty-guided refinement and adaptive distillation. Comprehensive experiments on three benchmark datasets achieve superior results, demonstrating that introducing the semi-supervision into FSCIL by our framework can achieve prominent and robust enhancement. 

In spite of the remarkable performance achieved by the UaD-CE framework, some aspects still need to be improved: (1) In each incremental learning session, the best proportion of unlabeled data in the semi-supervised learning process varies for different datasets. This proportion can affect not only the learning status of the current session but also the memorization of old knowledge. In the future, we will explore the root causes that affec the best proportion of labeled data. (2) The trade-off issue between old and new classes is challenging in distillation-based methods for FSCIL, and the adaptive weight of distillation loss is proposed to tackle this issue. More valid solutions will be explored in our future work. (3) Compared with pseudo-labeling-based methods, consistency regularization-based semi-supervised learning methods have achieved dominant performance. For future work, we will discuss the compatibility issue of this series of semi-supervised learning methods and distillation-based incremental learning framework.

\ifCLASSOPTIONcaptionsoff
  \newpage
\fi

\normalem
\bibliographystyle{IEEEtran}
\bibliography{references}

\end{document}